%% file: main.tex
\documentclass[runningheads]{llncs}

\usepackage{eccv}

\usepackage{eccvabbrv}

\usepackage{graphicx}
\usepackage{booktabs}

\usepackage[accsupp]{axessibility}  

\usepackage{hyperref}

\usepackage{orcidlink}

\usepackage{comment}
\usepackage{bbm}
\usepackage{booktabs, makecell}
\usepackage{arydshln}
\usepackage{mathtools}
\usepackage{adjustbox}

\usepackage{algorithm}
\usepackage{algpseudocode}
\usepackage{amsmath}
\usepackage{amssymb}
\providecommand\given{}

\newcommand\SetSymbol[1][]{
    \nonscript\:#1\vert
    \allowbreak
    \nonscript\:
    \mathopen{}}

\DeclarePairedDelimiterX\Set[1]\{\}{
    \renewcommand\given{\SetSymbol[\delimsize]}
    #1
}

\newcommand{\Softmaxsampler}[2]{S_{#1}(#2)}
\usepackage{amsmath}

\usepackage{kotex}

\usepackage{tcolorbox}
\tcbuselibrary{listings,breakable}

\newtcblisting{myverb}{
  breakable,
  listing only,
  colback=black!2,
  colframe=black!50,
  listing options={
    basicstyle=\fontsize{7}{8}\ttfamily\selectfont,
    breaklines=true,
    escapeinside={(*@}{@*)},
    literate={<}{\textless{}}1 {>}{\textgreater{}}1 {–}{--}1
             {_}{\_}1
  }
}

\hypersetup{
    colorlinks=true,
    filecolor=cyan,      
    urlcolor=magenta,
}

\begin{document}

\title{TMI: Text-to-Image Meets Image-to-Image for Complementary Data Synthesis to Boost Long-Tailed Instance Segmentation}

\titlerunning{TMI}

\author{Hyeonseop Song${}^{*}$\and
Seokhun Choi${}^{*}$\and
Hoseok Do${}^{\dag}$}

\authorrunning{H.~Song \& S.~Choi et al.}

\institute{
AI Lab, CTO Division, LG Electronics, Republic of Korea${}^{}$ \\
\email{\{hyeonseop.song, seokhun.choi, hoseok.do\}@lge.com}  \\
Project page: \href{https://seokhunchoi.github.io/TMI}{https://seokhunchoi.github.io/TMI} \\
}
\maketitle
\let\thefootnote\relax
\footnotetext{\textsuperscript{*}Equal contribution. \quad
\textsuperscript{\dag}Corresponding author.}
\begin{abstract}
Large-vocabulary instance segmentation is constrained by long-tailed category distributions and fine-grained inter-class ambiguity.
While data synthesis offers a promising alternative, current paradigms have complementary limitations: text-to-image (T2I) methods inherit noisy pseudo-labels and struggle on rare classes, whereas copy-paste methods compromise contextual realism.
To address these issues, we propose a hybrid pipeline coupling T2I generation with context-aware image-to-image (I2I) editing.
The T2I branch provides broad category and scene diversity, while a teacher-student scheme ensures label reliability by selectively retaining only prompt-specified categories.
To strengthen supervision for rare classes, we introduce VRAIN (Verified Rare-class Augmentation via INstructed editing), a novel I2I editor. VRAIN inserts high-confidence instances at semantically appropriate locations within in-the-wild scenes, yielding semantically coherent and visually natural edits that reduce domain gaps and enable targeted augmentation.
On the LVIS benchmark, our method surpasses existing baselines, improving overall AP by up to +4.0 points and rare-class AP by up to +9.5 points, while scaling effectively with backbone capacity.
\end{abstract}

\section{Introduction}
Instance segmentation is a critical task, underpinning diverse applications from autonomous driving~\cite{8014800, Cordts2016Cityscapes} and robotics~\cite{kimhi2025robot, zhang2025zisvfm, xie2022rice} to visual understanding~\cite{MOLINA2025129584, Wu_2025_CVPR, 10.1007/978-3-031-73195-2_14} and image editing~\cite{Liu_2024_CVPR, wang2024instancediffusion, Luo_2023_CVPR}.
However, achieving robust performance relies on large, densely annotated datasets, which are costly to build. This bottleneck is particularly acute for large-vocabulary benchmarks such as LVIS~\cite{gupta2019lvis} with their long-tailed category distributions.
While methods like re-weighting~\cite{tan2020equalization, tan2021equalization, 10094303}, balanced sampling~\cite{chang2021image, he2022relieving, 10094303}, and classifier calibration~\cite{wang2019classification, wang2020devil_simcal, pan2021model} help alleviate the imbalance, they cannot resolve the underlying data scarcity of rare categories~\cite{gupta2019lvis}.

\begin{figure*}[t]
\begin{center}
\includegraphics[width=0.9\linewidth]{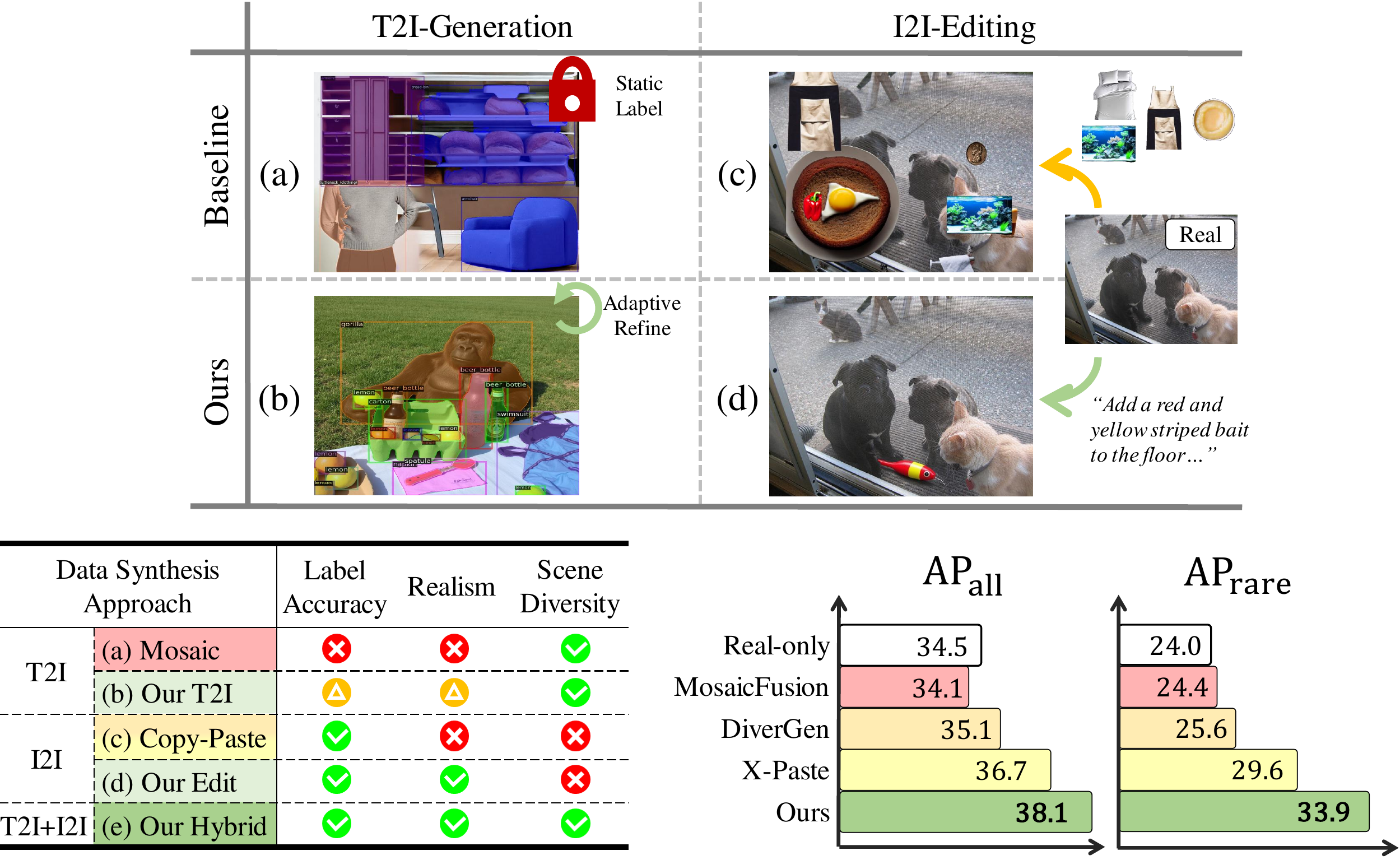}
\vskip -0.1in
\caption{\textbf{Mutually Complementary T2I-I2I Data Synthesis.}
\textbf{(a)} Existing T2I methods (\eg, MosaicFusion~\cite{xie2025mosaicfusion}) offer diverse scenes but lack label accuracy, while \textbf{(c)} Copy-Paste-based I2I methods (\eg, X-Paste~\cite{zhao2023xpaste} and DiverGen~\cite{fan2024divergen}) offer accurate labels but reduce realism. 
\textbf{(e)} Our hybrid framework combines two proposed modules: \textbf{(b)} a T2I branch for scene diversity with reliable labels via teacher-student adaptive pseudo-labeling, and \textbf{(d)} context-aware I2I edits for realistic, accurate instance-level supervision. 
This unified approach achieves state-of-the-art results on LVIS benchmark.
}
\label{fig:teaser}
\end{center}
\vspace{-25px}
\end{figure*}

This data scarcity has motivated recent work on synthetic data generation, leveraging advances in image generative modeling~\cite{ho2020denoising, rombach2022high, zhang2023adding_controlnet, flux2024} as a scalable alternative to manual annotation for improving long-tailed instance segmentation.
Current generative data synthesis methods fall into three paradigms: text-to-image (T2I), label-to-image (L2I), and image-to-image (I2I).
T2I methods~\cite{nguyen2023datasetdiffusion, wu2023datasetdm, xie2025mosaicfusion}, which first generate images from textual prompts and then assign semantic labels via pseudo-labeling, can introduce label noise, particularly for rare or fine-grained categories. 
L2I approaches~\cite{li2024aldm, ye2024seggen, yang2023freemask} synthesize images conditioned on segmentation masks, enabling precise layout control via mask-to-image modules such as ControlNet~\cite{zhang2023adding_controlnet}.
However, scaling this approach to large category sets remains challenging, often leading to label-image mismatches and inaccurate annotations.
In contrast, I2I approaches~\cite{zhao2023xpaste, fan2024divergen, loiseau2024reliability_genval} augment real datasets by adding new objects via copy-paste or inpainting, offering accurate annotations for added instances.
However, existing paste-based methods often suffer from domain gaps (\eg, illumination mismatches), while inpainting struggles with plausible mask placement.
Interestingly, when paste-based approaches explicitly target rare classes, performance on those classes often decreases rather than improves~\cite{fan2024divergen}.
This counterintuitive result suggests that naive pasting often harms contextual realism, leading the segmentation model to overfit to synthetic artifacts rather than learning true representations.

To overcome these limitations, we propose a hybrid data synthesis framework that strategically integrates T2I generation with context-aware I2I editing as illustrated in Fig.~\ref{fig:teaser}.
Our T2I branch generates diverse images of broad categories and scenes, from which pseudo labels are obtained through an offline labeler. 
Since offline pseudo labels are prone to noise due to the domain gap between real and generated images, we additionally employ a teacher-student scheme~\cite{tarvainen2017mean} that adapts to the T2I domain and progressively refines pseudo-label quality.
However, T2I generation alone is insufficient for rare classes, which remain difficult to label reliably.
We address this limitation with our novel I2I branch, \textbf{VRAIN} (\textbf{V}erified \textbf{R}are-class \textbf{A}ugmentation via \textbf{IN}structed editing), to enhance rare-class representation.
VRAIN operates as a two-stage pipeline: (\romannumeral 1) it first performs context-aware rare-class placement via instruction-based editing to ensure natural scene integration;
(\romannumeral 2) it then verifies the edit's semantic consistency and visual fidelity, generating a precise, high-fidelity annotation for the successfully integrated instance.
This two-stage ``place-and-verify'' process directly mitigates the contextual inconsistency of copy-paste methods and the mask-placement ambiguity inherent in inpainting-based approaches.
By yielding semantically coherent and visually natural edits, this process alleviates domain gaps and compositional artifacts, while also enabling targeted augmentation for rare categories.

Experimentally, our approach achieves state-of-the-art results on the LVIS benchmark, significantly outperforming existing paste-based augmentation and T2I baselines.
Notably, it demonstrates remarkable effectiveness for both overall and underrepresented categories.
Furthermore, our method scales effectively with larger backbones, demonstrating its practicality as a scalable solution for data generation in large-vocabulary instance segmentation.

In summary, our key contributions are as follows:
\begin{itemize}
\item We propose a unified T2I-I2I data synthesis framework that leverages T2I for scene diversity and I2I editing for enhanced realism and rare-class representation.
\item We design VRAIN, a novel ``place-and-verify'' I2I pipeline, which leverages instruction-based editing for semantically coherent placement and introduces a verification stage to generate accurate annotations.
\item We achieve state-of-the-art performance on the LVIS benchmark, especially on rare categories and demonstrate effective scaling with larger backbone models.
\end{itemize}

\section{Related Work}
\subsection{Image Generative Models}

High-quality T2I generation is dominated by diffusion models, which have rapidly evolved from DDPM~\cite{ho2020denoising} and latent diffusion~\cite{rombach2022high} to more efficient flow-matching formulations~\cite{lipman2022flow, liu2022flow}.
Modern architectures such as Flux~\cite{flux2024} and Qwen-Image~\cite{wu2025qwen} further boost fidelity and scalability by adopting Diffusion Transformer (DiT)~\cite{peebles2023scalable_dit} backbones with flow-matching objectives.
These T2I advances also empower instruction-based I2I editing---pioneered by InstructPix2Pix~\cite{brooks2023instructpix2pix} and extended by unified frameworks like Flux-Kontext~\cite{labs2025fluxkontext} and Qwen-Image-Edit~\cite{wu2025qwen}---enabling semantically consistent, language-guided generation and editing.
Our dataset synthesis framework is built on these generative and editing developments.

\subsection{Data Generation for Segmentation Task}
Recent advances in generative modeling have led to various data generation strategies for segmentation, broadly categorized as T2I, L2I, and I2I.
\textbf{T2I methods} generate images from textual prompts, often using diffusion attention maps~\cite{nguyen2023datasetdiffusion, wu2023diffumask, xie2025mosaicfusion} or fine-tuning perception decoders~\cite{wu2023datasetdm} for pseudo-labeling in few-shot settings.
For example, MosaicFusion~\cite{xie2025mosaicfusion} divides the image canvas into mosaic~\cite{bochkovskiy2020yolov4}-like regions and simultaneously generates multiple objects.
However, most T2I approaches are effective on small-scale data or for a limited number of categories~\cite{nguyen2023datasetdiffusion, wu2023diffumask}, struggling to scale to large-scale, in-the-wild scenarios.

\textbf{L2I methods} generate images conditioned on segmentation masks, enabling precise control over object layout.
Some works~\cite{li2024aldm, yang2023freemask} utilize mask-to-image modules like ControlNet~\cite{zhang2023adding_controlnet, xue2023freestylenet}, while others~\cite{ye2024seggen} extend this framework by designing text-to-mask modules for mask diversity.
However, extending L2I to large-vocabulary datasets such as LVIS~\cite{gupta2019lvis} remains challenging.
Mask-only conditioning often fails to faithfully render the semantics of numerous fine-grained categories, leading to label-image mismatches.

\textbf{I2I methods} expand datasets by editing existing images.
Copy-paste techniques, such as X-Paste~\cite{zhao2023xpaste} and DiverGen~\cite{fan2024divergen}, leverage generative models to create diverse instance pools for composition onto real images.
This successfully covers large categories where L2I methods struggle; however, the pasted instances often appear visually distinct from their surroundings (\eg, illumination or blending mismatches), limiting training gains.
Inpainting-based approaches~\cite{loiseau2024reliability_genval, de2024placing_poc} synthesize objects into masked regions. 
While effective in constrained domains (\eg, road scenes~\cite{loiseau2024reliability_genval, de2024placing_poc}), they have not scaled to in-the-wild, large-category datasets like LVIS, due to the challenges of realistic mask placement and high semantic diversity of scenes.

Our hybrid framework addresses these complementary failures.
We harness T2I for diversity but mitigate its label noise by employing prompt-consistent filtering and teacher-student refinement.
We then use our I2I branch to address both T2I's rare-class weakness and prior I2I's realism gap, providing high-fidelity, targeted augmentation that scales effectively on LVIS.

\section{Methods}
\begin{figure*}[t]
\begin{center}
\includegraphics[width=\textwidth]{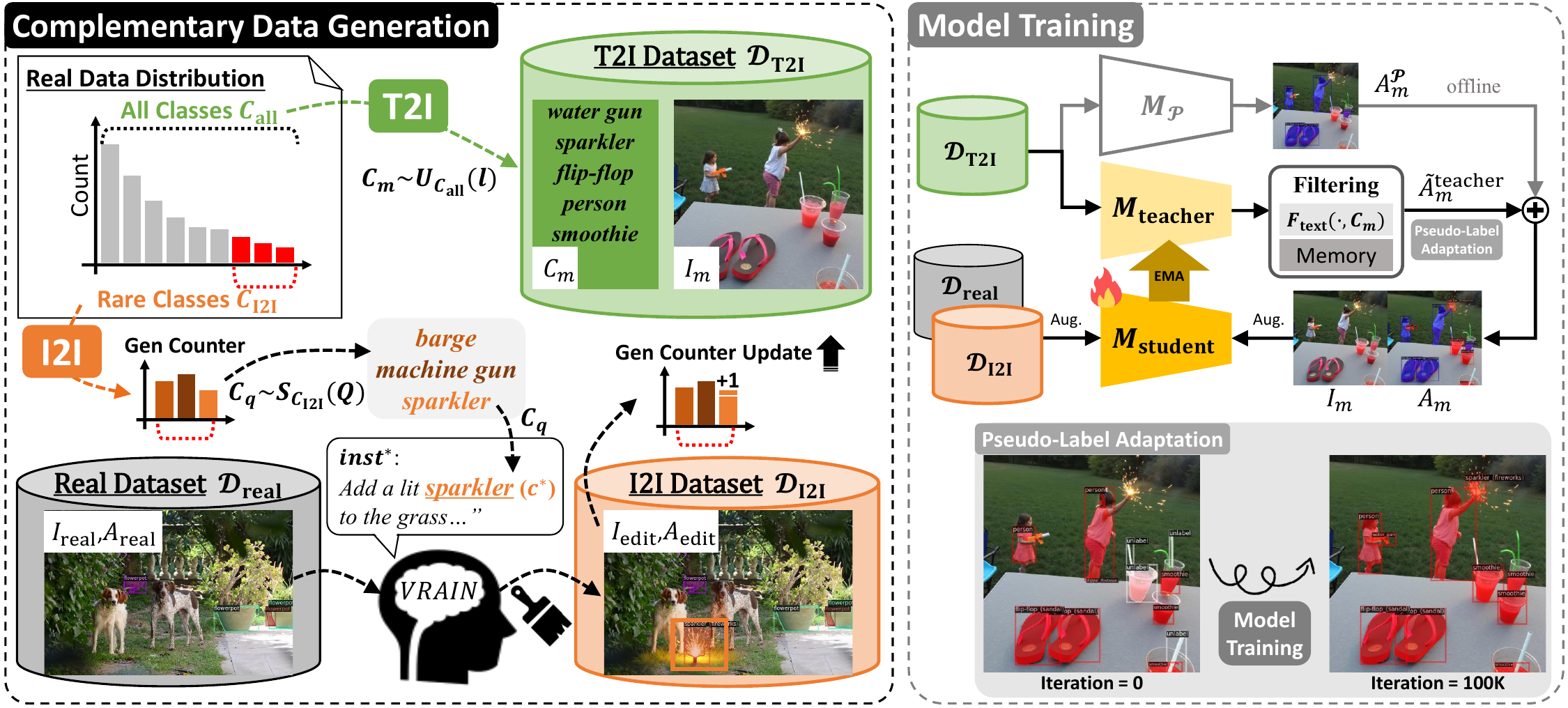}
\vskip -0.1in
\caption{\textbf{Overview of Hybrid Data Generation and Training Framework.}
We combine two complementary paradigms: (i) T2I generation for diverse scenes over all categories $C_\text{all}$ and (ii) I2I editing (VRAIN) for rare categories $C_\text{I2I}$ with accurate instance-level supervision.
(iii) A student $M_\text{student}$ jointly learns from both $\mathcal{D}_{\text{T2I}}$ and $\mathcal{D}_{\text{I2I}}$. 
For $\mathcal{D}_{\text{T2I}}$ images $I_m$, an EMA teacher $M_\text{teacher}$ generates refined pseudo-labels $\tilde{A}_m^\text{teacher}$, merged with offline labels ${A}_{m}^\mathcal{P}$ to form the final supervision $A_m$.
As training proceeds, the teacher adapts to the T2I domain, progressively improving initially noisy pseudo-labels (pseudo-label adaptation).
For example, rare classes (\eg, \textit{sparkler} and \textit{water gun})---missed by the offline labeler and initially unlabeled---are gradually labeled correctly as the teacher adapts with the support from rare-class-targeted I2I data.
}
\label{fig:method_overview}
\end{center}
\vspace{-7mm}
\end{figure*}

We generate synthetic training data through two complementary paradigms: (\romannumeral 1) T2I generation for diverse scene synthesis, and (\romannumeral 2) I2I editing for context-aware augmentation of real images.
The T2I branch (Sec.~\ref{sec:t2i}) provides broad semantic coverage across all categories in the long-tailed dataset. 
However, assigning accurate annotations remains a challenge; conventional offline models, trained exclusively on real images, produce noisy pseudo-labels due to the domain gap between real and generated images.
We therefore treat this T2I set as a source of weakly-labeled data, $\mathcal{D}_{\text{T2I}}$.
In contrast, our I2I branch (VRAIN, Sec.~\ref{sec:i2i}) enhances rare-class representation. 
Because these edits are driven by specific instructions, the identity and location of the newly inserted instance are explicitly known, yielding accurate, instance-level labeled data ($\mathcal{D}_{\text{I2I}}$) that provide reliable supervision for rare categories where T2I struggles.
Our final synthesized dataset is the union of these two complementary sets, $\mathcal{D}_{\text{gen}} = \{{\mathcal{D}_{\text{T2I}}, \mathcal{D}_{\text{I2I}}}\}$.
Finally, we introduce a teacher-student scheme (Sec.~\ref{sub:3_3_model_training}) to train the student model on this hybrid dataset.
In this semi-supervised loop, an EMA-updated teacher generates dynamic, online pseudo-labels for the $\mathcal{D}_{\text{T2I}}$.
The student is jointly trained on these progressively refined T2I pseudo-labels to expand category and scene diversity, while also incorporating the labeled I2I data for accurate, instance-level supervision. 
An overview of the entire process is depicted in Fig.~\ref{fig:method_overview}.

\subsection{Text-to-Image Generation}
\label{sec:t2i}
Our T2I method consists of two stages: (\romannumeral 1)~generating a set of text prompts and corresponding images, and (\romannumeral 2)~offline pseudo-labeling with text-consistent filtering.
First, to generate text prompt set $\mathcal{T}$ of size $N_\text{text}$, we randomly sample a subset $C_m$ of size $l$ from the total category set $C_\text{all}$. We then create a descriptive prompt $t_m$ composed of these selected classes using GPT-4o~\cite{hurst2024gpt4o}:
\begin{equation}
\label{eq:T2I_text_generation}
\mathcal{T} =
\Set*{
    t_{m} \;\given\;
    t_{m} = \mathrm{GPT\text{-}4o}(C_{m})
}.
\end{equation}
Subsequently, we generate an image $I_{m}$ from the created text prompt using the image generation model $\Phi_{\text{T2I}}$~\cite{flux2024} and construct $N_{\text{T2I}}$ unlabeled T2I dataset as image-category pairs: $\mathcal{D}_{\text{T2I}}= \{(I_{m},C_{m}) \mid I_{m}=\Phi_{\text{T2I}}(t_{m})\}$

\begin{table}[t]
  \centering
  \vspace{2mm}
  \resizebox{0.65\columnwidth}{!}
  {%
  \begin{tabular}{l c c c c}
    \toprule
    \textbf{Method} &
      {\textbf{AP$^{\text{box}}$}} &
      {\textbf{AP$^{\text{mask}}$}} &
      {\textbf{AP$^{\text{box}}_{\mathrm{r}}$}} &
      {\textbf{AP$^{\text{mask}}_{\mathrm{r}}$}} \\
    \midrule
    (a) Real-only                                     & 34.5 & 30.8 & 24.0 & 21.6 \\ \hdashline
    (b) High-threshold Filtering                      & 35.8 & 32.0 & 31.2 & 28.0 \\
    (c) Text-consistent Filtering                     & \textbf{36.7} & \textbf{33.0} & \textbf{32.1} & \textbf{28.9} \\
    \bottomrule
  \end{tabular}}
  \caption{\textbf{Comparison of T2I Pseudo-label Filtering Strategies.}
We compare models trained with T2I data, where the raw predictions $\hat{A}_{m}^\mathcal{P}$ are filtered using either (b) a high-threshold strategy or (c) our text-consistent strategy.
Our method (c) retains semantically valid, low-confidence instances that high-threshold filtering discards, resulting in improved overall and rare-class AP.}
  \label{tab:ablation_text_filter}
\vspace{-9mm}
\end{table}

Second, we pre-compute a set of offline labels for this dataset.
We process each image $I_m$ with a pre-trained public instance segmentation model~\cite{zong2023detrs} $M_\mathcal{P}$ to obtain an initial set of raw predictions, $\hat{A}_{m}^\mathcal{P} = M_{\mathcal{P}}(I_{m})$.
To leverage the prior knowledge that $I_{m}$ was generated from the prompt $t_{m}$ containing classes $C_{m}$, we apply a text-consistent filtering process $F_{\text{text}}$. For the $k$-th predicted instance annotation $a_k^\mathcal{P}$, this filter retains only the instances whose predicted classes belong to $C_{m}$:
\begin{equation}
\label{eq:text_filter}
{A}_{m}^\mathcal{P} = F_{\text{text}}(\hat{A}_{m}^\mathcal{P},C_{m}) = \Set*{a_k^\mathcal{P} \mid a_k^\mathcal{P} \in \hat{A}_{m}^\mathcal{P},\; \text{class}(a_k^\mathcal{P}) \in C_{m}}.
\end{equation}
The effectiveness of this text-consistent filtering is validated in Table~\ref{tab:ablation_text_filter}.
Our filtering method yields superior results compared to high-threshold instance mining. This demonstrates that our filtering strategy, which leverages prompt knowledge, is more effective than relying solely on high-confidence scores.
It successfully retains valid instances, even those with low-confidence, that are consistent with the text prompt.
These filtered offline labels, ${A}_{m}^\mathcal{P}$, are then used as one of the supervision sources in our training pipeline, detailed in Sec.~\ref{sub:3_3_model_training}.

\begin{figure*}[t]
\begin{center}
\includegraphics[width=0.95\linewidth]{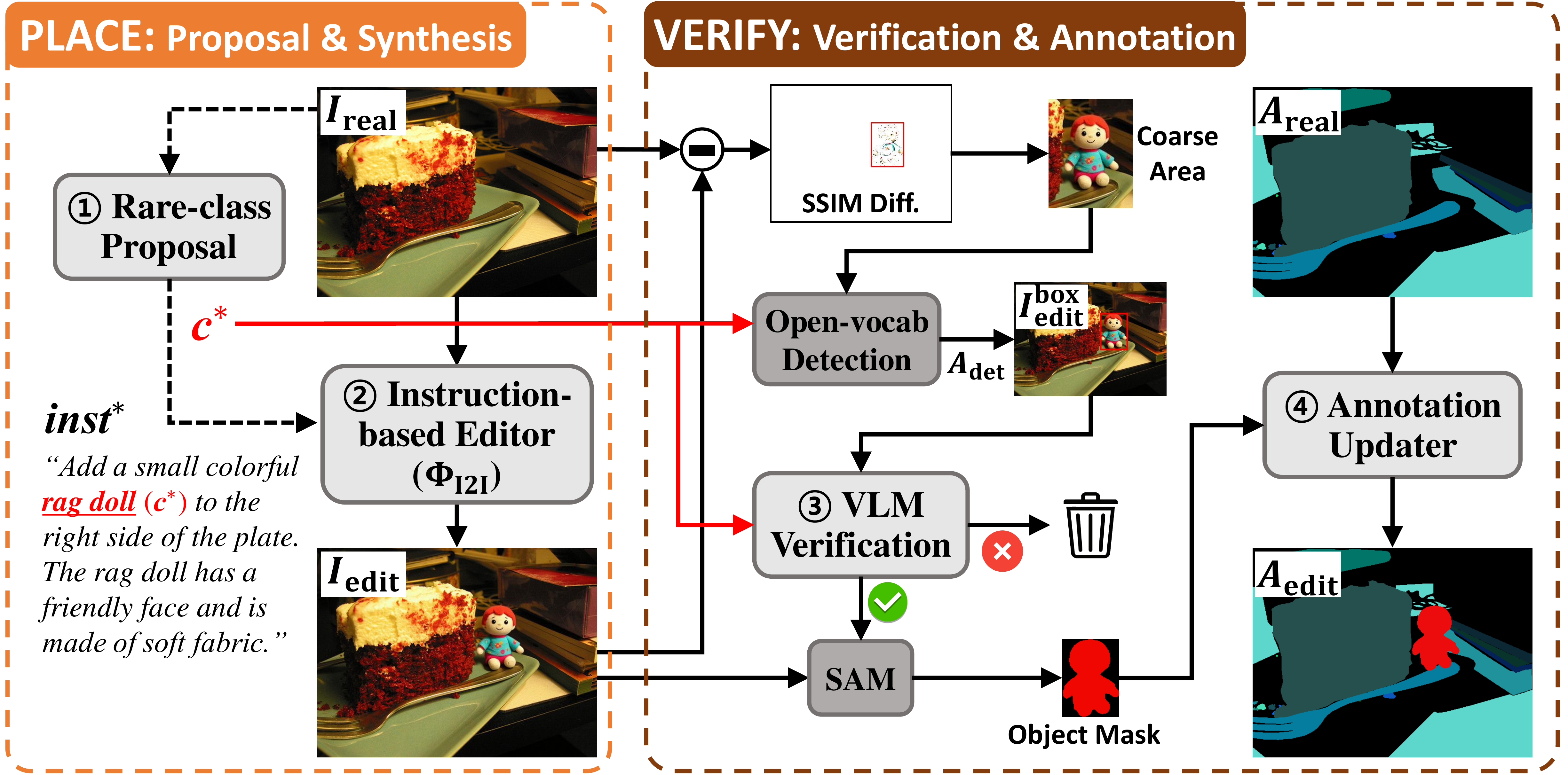}
\vskip -0.1in
\caption{
\textbf{VRAIN ``Place-and-Verify'' Pipeline.}
Our two-stage framework ensures high-fidelity I2I editing. \textbf{(i) Place:} A VLM proposes a semantically coherent instruction (${inst}^*$) for inserting a rare class ($c^*$) that fits naturally within $I_\text{real}$. 
An instruction-based editor $\Phi_\text{I2I}$ then synthesizes $I_\text{edit}$.
\textbf{(ii) Verify:} $I_\text{edit}$ is then validated. The new instance is localized via SSIM difference and open-vocab detection, semantically confirmed by a VLM filter, and masked using SAM.
Finally, an annotation updater resolves occlusions with $A_\text{real}$, integrating the new instance to produce the trustworthy final annotation $A_\text{edit}$.
}
\label{fig:I2I}
\end{center}
\vspace{-25px}
\end{figure*}

\subsection{Image-to-Image Editing}
\label{sec:i2i}
While our T2I branch provides broad diversity, it struggles to produce accurate instance-level annotations, especially for rare categories prone to label noise.
A high-quality seed set of verified rare-class annotations is therefore essential, as it bootstraps the online pseudo-labeling methods (\eg., EMA teacher~\cite{tarvainen2017mean}). 
I2I approaches, which can add new objects with accurate annotations, offer a promising solution.
However, prior I2I methods suffer from critical flaws: copy-paste techniques produce contextually inconsistent artifacts, while inpainting-based methods struggle to identify suitable mask regions in complex scenes. Motivated by these limitations, we propose VRAIN (\textbf{V}erified \textbf{R}are-class \textbf{A}ugmentation via \textbf{IN}structed editing), illustrated in Fig.~\ref{fig:I2I}.
VRAIN is a principled two-stage pipeline designed to overcome these realism and placement challenges, performing (1) context-aware rare-class placement via instruction-based editing, and (2) verification of the edit's fidelity to generate trustworthy annotation.
\subsubsection{Rare-class Proposal and Synthesis.}
A central challenge in rare-class generation is selecting a semantically fitting category and determining its placement, ensuring the inserted instances are contextually coherent with the scene. This process is key to reducing the unrealistic compositions of previous copy-paste methods.
To this end, we leverage a Vision Language Model (VLM)~\cite{zhu2025internvl3} to ensure semantic consistency, by selecting the most suitable category from a candidate set for a given image, and then generating its placement instruction.
This candidate set ($C_q$) is formed by sampling $Q$ categories from all target categories $C_\text{I2I}$ for a given real image $I_\text{real}$ from the training dataset $\mathcal{D}_\text{real}=\Set{({I_\text{real}}, {A_\text{real}})}$.
The sampling is performed using a softmax-based sampler ($C_q \sim \Softmaxsampler{C_\text{I2I}}{Q}$) that is weighted by the generation count of each class (initialized to zero), penalizing frequently generated classes.
The VLM is then prompted with $I_\text{real}$ and the candidate set $C_q$ to recommend a suitable category $c^*$ along with the corresponding textual instruction $inst^*$:
\begin{equation}
\label{eq:recommendation_module}
(c^*,\; inst^*) = \mathrm{VLM}(I_\text{real},\; C_q),
\end{equation}
where $inst^*$ is the natural language instruction fed to the editor, while $c^*$ is the explicit category label passed to the annotation stage (see the supplementary material for details).
Finally, $inst^*$ is used by an instruction-based editing model $\Phi_{\text{I2I}}$~\cite{labs2025fluxkontext} to produce the edited image: $I_{\text{edit}} = \Phi_{\text{I2I}}(I_\text{real}, inst^*)$.

\subsubsection{Verification and Annotation.}
Although the instruction-based editor $\Phi_{\text{I2I}}$ produces visually plausible results, these edits may suffer from contextual mismatches or rendering artifacts.
To ensure annotation quality, VRAIN performs a verification and annotation stage that evaluates both semantic consistency and visual fidelity of each edited image.
First, to locate the edited region, we compute a structural similarity (SSIM~\cite{ssim}) difference map between $I_{\text{real}}$ and $I_{\text{edit}}$, thresholding at $\tau_{\text{edit}}$ to identify modified pixels.
Second, to detect the new object, we apply an open-vocabulary detector~\cite{fu2025llmdet} within this coarse area, searching for the intended category $c^*$ and producing a set of candidate detections $A_{\text{det}}$.

However, $A_{\text{det}}$ may contain false-positives or artifacts. To semantically verify these candidates, we introduce a VLM-based filter, $F_{\text{VLM}}(A_{\text{det}}, c^*)$.
For each detection $a_k\in\!A_{\text{det}}$, we prepare an image $I_{\text{edit}}^{\text{box}}(a_k)$ by overlaying its bounding box in red on the edited image.
The VLM is then prompted with this visualization along with a template-based question $t^{c^*}$ (\eg, ``Is the red bounding box in the image a $c^*$?'').
The VLM outputs a binary `Yes' or `No' response.
A detection $a_k$ is confirmed only if the VLM responds `Yes':
\begin{equation}
\label{eq:vlm_filter}
F_{\text{VLM}}(A_{\text{det}}, c^*) =
\Set*{ a_k \mid a_k\!\in\!A_{\text{det}},
\mathrm{VLM}(I^{\text{box}}_{\text{edit}}, t^{c^*}; a_k)\!=\text{Yes}}.
\end{equation}
For all verified objects in $F_{\text{VLM}}$, we obtain instance masks utilizing Segment Anything Model (SAM)~\cite{ravi2024sam2}.
These new masks are then added to the original annotation $A_\text{real}$.
However, this integration can create occlusions, where new SAM masks overlap with an existing ground-truth mask from $A_\text{real}$. 
Thus, we update in $A_\text{real}$ by subtracting the new SAM masks from any overlapping original masks to produce the final annotation $A_\text{edit}$. 
If this process yields at least one verified instance mask, the generation count for category $c^*$ is incremented by one.
This entire procedure is repeated until the total count for $C_\text{I2I}$ reaches the target $N_\text{I2I}$, resulting in the final dataset $\mathcal{D}_\text{I2I}=\Set{({I_\text{edit}}, {A_\text{edit}})}$ of size $N_\text{I2I}$.

\subsection{Model Training}
\label{sub:3_3_model_training}
The straightforward way to train an instance segmentation model~\cite{zhou2021probabilistic_ceternet2} using the two complementary synthesized datasets, 
$\mathcal{D}_{\text{gen}} = \{\mathcal{D}_{\text{T2I}}, \mathcal{D}_{\text{I2I}}\}$, is to combine them with the real dataset $\mathcal{D}_{\text{real}}$ for joint training.
However, unlike $\mathcal{D}_{\text{real}}$ or $\mathcal{D}_{\text{I2I}}$, all images in $\mathcal{D}_{\text{T2I}}$ are fully generated, which can lead to domain gaps and unreliable pseudo-labels using a static labeler trained only on real data.
To address this issue, we adopt an EMA-based teacher-student scheme~\cite{tarvainen2017mean}, where the student model $M_\text{student}$ learns from supervision signals generated by the teacher model $M_\text{teacher}$. Its parameters are updated as $\Theta_\text{teacher} \leftarrow \gamma \Theta_\text{teacher} + (1 - \gamma) \Theta_\text{student}$, where $\Theta_\text{teacher}$ and $\Theta_\text{student}$ denote the parameters of the teacher and student models, respectively, and $\gamma$ is the EMA decay rate.
The teacher gradually adapts to the T2I domain during training, thereby producing more accurate pseudo-labels for T2I data.
Moreover, because the teacher is also trained on the high-quality, rare-class-focused $\mathcal{D}_{\text{I2I}}$, it becomes more adept at assigning more precise labels even for rare categories in $\mathcal{D}_{\text{T2I}}$.

Specifically, at each training step, a batch is sampled from $\{{\mathcal{D}_{\text{real}}, \mathcal{D}_{\text{I2I}}, \mathcal{D}_{\text{T2I}}}\}$.
For images in $\mathcal{D}_{\text{real}}$ and $\mathcal{D}_{\text{I2I}}$, the student is trained directly using their ground-truth annotations ($A_\text{real}$ and $A_\text{edit}$).
On the other hand, for images in $\mathcal{D}_{\text{T2I}}$, the student is trained on hybrid pseudo-labels created by processing and merging the offline ($A_m^\mathcal{P}$) and new online ($A_m^\text{teacher}$) labels generated by the teacher.

This training-time process for online labels is as follows. 
First, the online EMA-based teacher $M_\text{teacher}$ produces raw predictions $\hat{A}_{m}^\text{teacher}$, which are then split into two groups according to their predicted class $\hat{c}$ and the sampled category set $C_m$.
Prompt-consistent instances ($\hat{c} \in C_m$) are filtered using a class-wise adaptive threshold. This threshold is dynamic; we maintain a per-category running memory of recent confidences, and the threshold for a given class is set to $\tau_{label}$ times the mean confidence of that memory, allowing the filter to adapt as the teacher improves.
Prompt-inconsistent instances ($\hat{c} \notin C_m$) are assigned an `unlabeled' class if they are detected with high confidence above $\tau_{unlabel}$ times the mean confidence.
These `unlabeled' instances are excluded only from the classification loss but are still included in box and mask losses~\cite{zhou2021probabilistic_ceternet2}, providing implicit localization supervision from objects that were confidently detected but not explicitly specified in the prompt.
These two sets of processed labels (the adaptively-thresholded and the unlabeled instances) together form the complete online pseudo-label set, $\tilde{A}_m^\text{teacher}$.
Finally, we merge this dynamic online set $\tilde{A}_m^\text{teacher}$ with the static offline pseudo-labels ${A}_{m}^\mathcal{P}$ via IoU-based matching to produce the student's final supervision, $A_m$. This strategy combines their complementary strengths: the offline set ${A}_{m}^\mathcal{P}$ provides a stable, precision-oriented baseline of high-confidence instances with conservative coverage of common categories, while the online set $\tilde{A}_m^\text{teacher}$ introduces progressively refined, domain-adapted labels, especially for rare classes thanks to the auxiliary guidance from I2I data. The student $M_\text{student}$ thus learns from both a robust baseline and dynamic, in-domain corrections.

\section{Experiments}
\subsection{Experimental Settings}
\subsubsection{Dataset.}
We conducted experiments on LVIS~\cite{gupta2019lvis}, a large-scale instance segmentation dataset featuring a long-tailed distribution of 1,203 categories.
It contains 100K training and 20K validation images with 2M annotations.
Based on the number of training images per category, categories are divided into three frequency groups: rare (1--10 images; 337 categories), common (11--100 images; 461 categories), and frequent ($>$100 images; 405 categories).

\subsubsection{Baselines.}
We compare our method against three baselines: MosaicFusion~\cite{xie2025mosaicfusion} (a T2I method), X-Paste~\cite{zhao2023xpaste}, and DiverGen~\cite{fan2024divergen} (I2I copy-paste methods).
We use CenterNet2~\cite{zhou2021probabilistic_ceternet2}, a widely used model for large-vocabulary instance segmentation, for all our experiments, with two backbone configurations: ResNet-50~\cite{he2016deep} (640 resolution, batch size 32) and Swin-L~\cite{liu2021swin} (896 resolution, batch size 16).
Following the convention of prior work~\cite{zhao2023xpaste, fan2024divergen}, all models are trained for 180K iterations, initializing from their respective ``Real-only'' pretrained checkpoints.
To ensure a fair comparison under identical experimental settings, we reproduced DiverGen using their official code, generating 1K synthetic instance pool per category (1.2M in total) with two generative models, and also reproduced MosaicFusion following its original implementation.
For the evaluation metrics, we evaluate performance using LVIS box average precision $\text{AP}^\text{box}$ and mask average precision $\text{AP}^\text{mask}$. 
We also report rare-category results $\text{AP}^\text{box}_\text{r}$ and $\text{AP}^\text{mask}_\text{r}$.

\subsubsection{Implementation Details.}
We generated 200K T2I images ($N_\text{T2I}=200\text{K}$) using Flux.1-dev~\cite{flux2024} from a set of 40K prompts ($N_\text{text}=40\text{K}$), where each prompt was created by randomly sampling $l \in [5,10]$ categories from the total LVIS category set.
We also generated 80K I2I images ($N_\text{I2I}=80\text{K}$) from the LVIS training set using Flux.1 Kontext-dev~\cite{labs2025fluxkontext}, targeting rare LVIS categories with $Q=5$ and an edit-region threshold $\tau_{\text{edit}}=5$.
We set the EMA decay rate to $\gamma = 0.999$, the same as the baselines, but in our framework it is used to adaptively refine pseudo-labels rather than just for evaluation.
We set $\tau_{label}=0.7$ and $\tau_{unlabel}=1.2$ for dynamic threshold in online pseudo-labeling with running memory size 100 per class.
We used InternVL3-14B~\cite{zhu2025internvl3} as the VLM for VRAIN's recommendation and verification steps.
Further details on VLM implementation including prompt creation are provided in the supplementary material.

\begin{table*}[t]
  \centering
  \vspace{2mm}
  \resizebox{1.0\textwidth}{!}{%
  \begin{tabular}{c|l|cc|cccc}
    \toprule
    \textbf{Backbone} &
    \textbf{Method} &
    \textbf{T2I Dataset} &
    \textbf{I2I Dataset} &
    \textbf{AP$^{\text{box}}$} &
    \textbf{AP$^{\text{mask}}$}&
    \textbf{AP$_\mathrm{r}^{\text{box}}$} &
    \textbf{AP$_\mathrm{r}^{\text{mask}}$} \\
    \midrule
    \makecell{ResNet-50} &
    \makecell[l]{Real-only~\cite{zhou2021probabilistic_ceternet2}} &
    - & - &
    34.5 & 30.8 & 24.0 & 21.6 \\ \hdashline
    \makecell{ResNet-50} &
    \makecell[l]{MosaicFusion~\cite{xie2025mosaicfusion}} &
    \checkmark & &
    34.1 & 30.4 & 24.4 & 22.5  \\
    \makecell{ResNet-50} &
    \makecell[l]{DiverGen~\cite{fan2024divergen}} &
     & \checkmark &
    35.1 & 31.2 & 25.6 & 23.8 \\
    \makecell{ResNet-50} &
    \makecell[l]{X-Paste~\cite{zhao2023xpaste}} &
     & \checkmark &
    36.7 & 33.0 & 29.6 & 27.8 \\
    \makecell{ResNet-50} &
    \makecell[l]{\textbf{Ours}} &
    \checkmark &
    \checkmark &
    \textbf{38.1} & \textbf{34.0} & \textbf{33.9} & \textbf{31.7} \\
    \midrule
    \makecell{Swin-L} &
    \makecell[l]{Real-only~\cite{zhou2021probabilistic_ceternet2}} &
    - & - &
    47.5 & 42.3 & 41.4 & 36.8   \\ \hdashline
    \makecell{Swin-L} &
    \makecell[l]{MosaicFusion~\cite{xie2025mosaicfusion}} &
    \checkmark & &
    47.7 & 42.8 & 41.3 & 37.5  \\
    \makecell{Swin-L} &
    \makecell[l]{DiverGen~\cite{fan2024divergen}} &
     & \checkmark &
    49.6 & 44.2 & 44.5 & 39.8   \\
    \makecell{Swin-L} &
    \makecell[l]{X-Paste~\cite{zhao2023xpaste}} &
     & \checkmark &
    50.1 & 44.4 & 48.2 & 43.3  \\
    \makecell{Swin-L} &
    \makecell[l]{\textbf{Ours}} &
    \checkmark & \checkmark &
    \textbf{50.7} & \textbf{45.2} & \textbf{49.1} & \textbf{44.0}  \\
    \bottomrule
  \end{tabular}}
  \caption{
  \textbf{Comparison with Baselines on LVIS validation Dataset.}
Our hybrid T2I-I2I approach consistently outperforms all baselines: the Real-only model, the T2I-only method (MosaicFusion), and copy-paste I2I methods (DiverGen, X-Paste). This superior performance holds across both ResNet-50 and Swin-L backbones, delivering particularly strong gains on rare-class metrics and demonstrating the effectiveness of combining both data synthesis paradigms.
}
  \label{tab:main_table}
\vspace{-8mm}
\end{table*}

\begin{figure*}[t]
\begin{center}
\includegraphics[width=1.0\linewidth]{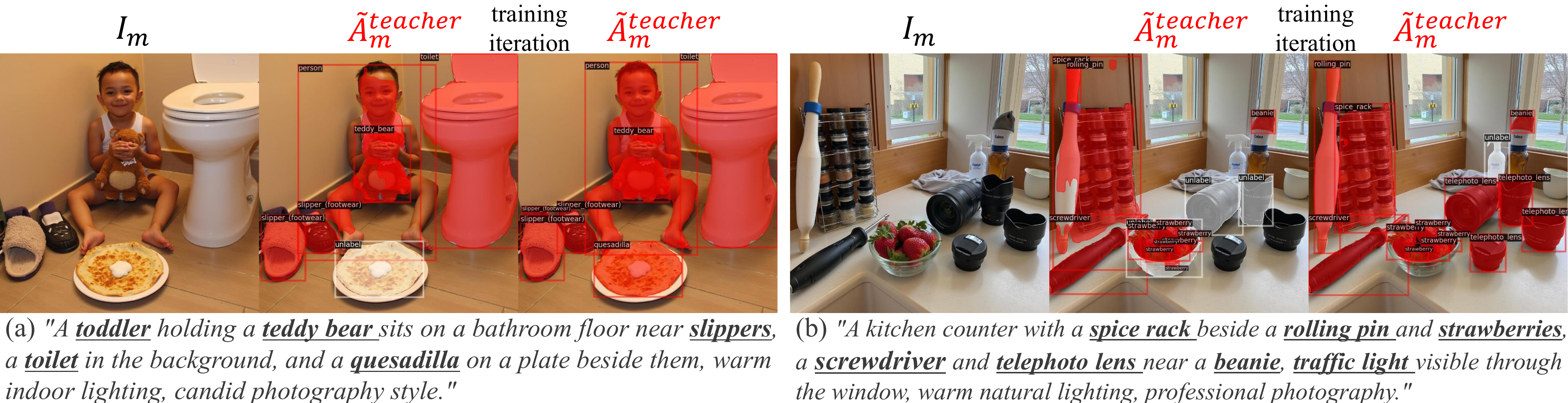}
\vskip -0.1in
\caption{
\textbf{Examples of Pseudo-label Adaptation in T2I images.}
In each T2I prompt, boldfaced and underlined words indicate the target categories for generation.
As the teacher model updates, it adapts to the T2I domain and enhances pseudo-label ($\tilde{A}_m^\text{teacher}$) quality---(a) improving the \textit{toddler} mask and labeling previously missed objects (\eg, \textit{quesadilla}), and (b) refining masks (\eg, \textit{spice rack}, \textit{rolling pin}) while newly labeling a \textit{telephoto lens}.
}
\label{fig:T2I_adaptation}
\end{center}
\vspace{-24px}
\end{figure*}

\subsection{Comparison}
We evaluated our approach against various data synthesis methods, as shown in Table~\ref{tab:main_table}.
MosaicFusion, a rare-class-targeted T2I method, offers only marginal gains over the real-only model, indicating that despite its scene diversity, the limited label accuracy of T2I generation constrains performance on long-tailed datasets.
Copy-paste I2I methods like DiverGen and X-Paste provide more accurate instance-level labels and achieve larger improvements, yet still lag behind our hybrid T2I-I2I framework.
Our method uniquely combines the strengths of both paradigms.
It leverages T2I for broad diversity like MosaicFusion, but uses our teacher-student adaptive pseudo-labeling to ensure high quality label.
By further incorporating VRAIN (I2I) to inject rare-class data with precise instance-level labels, it enhances performance on underrepresented categories and bootstraps online pseudo-labeling, boosting overall accuracy.
Fig.~\ref{fig:T2I_adaptation} demonstrates the EMA teacher's progressive improvement of pseudo-labels on T2I images during training---including the detection and mask refinement of previously missed rare-class objects such as \textit{quesadilla} and \textit{telephoto lens}---effectively bootstrapping the online pseudo-labeling process.
This synergy achieves the largest gains across all categories and especially for rare classes, improving AP$^\text{box}$ from 34.5 → 38.1 (ResNet-50) and 47.5 → 50.7 (Swin-L) overall, and AP$_r^\text{box}$ from 24.0 → 33.9 (ResNet-50) and 41.4 → 49.1 (Swin-L) for rare categories.

\begin{table}[t]
  \centering
  \vspace{2mm}
  \resizebox{0.6\columnwidth}{!}{%
  \begin{tabular}{l c c c c}
    \toprule
    \textbf{Method} &
      {\textbf{AP$^{\text{box}}$}} &
      {\textbf{AP$^{\text{mask}}$}} &
      {\textbf{AP$^{\text{box}}_{\mathrm{r}}$}} &
      {\textbf{AP$^{\text{mask}}_{\mathrm{r}}$}} \\
    \midrule
    (a) Real-only\hspace*{24mm} & 47.5 & 42.3 & 41.4 & 36.8 \\ \hdashline
    (b) MosaicFusion~\cite{xie2025mosaicfusion} & 47.7 & 42.8 & 41.3 & 37.5 \\
    (c) DiverGen (rare-target)~\cite{fan2024divergen} & 47.3 & 42.2 & 34.5 & 31.2 \\
    (d) $\mathcal{D}_\text{I2I}$ only (VRAIN) & \textbf{48.1} & \textbf{43.3} & \textbf{42.9} & \textbf{39.5} \\
    \bottomrule
  \end{tabular}}
  \caption{
  \textbf{Comparison with Rare-class-targeted Data Synthesis Approaches.}
Among methods generating data for rare categories, (c) DiverGen shows decreased rare-class performance due to reduced realism, whereas (d) our I2I approach effectively and practically improves rare-class performance with high visual fidelity, accurate labels.
}
  \label{tab:rare_target_table}
  \vspace{-9mm}
\end{table}

To further analyze rare-class-focused strategies, we compare approaches that synthesize data exclusively for rare categories, as shown in Table~\ref{tab:rare_target_table}.
Although designed for rare-class enhancement, MosaicFusion again provides a marginal improvement over the real-only model, indicating that the limited label reliability of T2I generation constrains its effectiveness.
Notably, the rare-class-focused variant of DiverGen catastrophically underperforms the real-only baseline on rare categories (AP$^\text{box}_\text{r}$ 41.4 → 34.5). This contradictory result highlights that naive copy-pasting harms performance due to its poor contextual coherence and realism, encouraging overfitting to pasted instances rather than learning true representations.
In contrast, our rare-class-focused I2I approach (VRAIN) successfully boosts rare-class performance, benefiting from high visual fidelity and precise instance-level labeling.
These results demonstrate the effectiveness and practical applicability of our method for improving underrepresented categories.

\begin{figure*}[t]
\begin{center}
\includegraphics[width=1.0\linewidth]{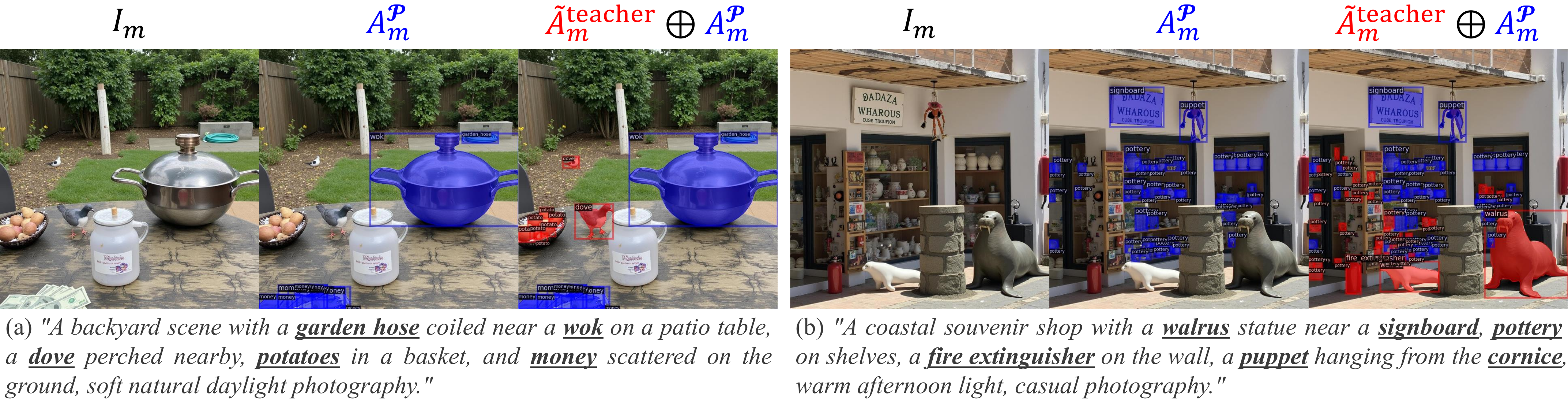}
\vskip -0.1in
\caption{
\textbf{Examples of Final Pseudo-labels for T2I images.}
In each T2I prompt, boldfaced and underlined words indicate the target categories for generation. 
Teacher-generated pseudo-labels (\textcolor{red}{red}) are merged ($\oplus$) with offline pseudo-labels (\textcolor{blue}{blue}) to form the final supervision. 
As the teacher adapts to the T2I domain, it successfully labels rare classes---(a) \textit{dove} and (b) \textit{walrus}.
}
\label{fig:T2I_merge}
\end{center}
\vspace{-20px}
\end{figure*}

\begin{table}[t]
  \centering
  \vspace{2mm}
  \resizebox{0.65\columnwidth}{!}{%
  \begin{tabular}{l c c c c}
    \toprule
    \textbf{Method} &
      {\textbf{AP$^{\text{box}}$}} &
      {\textbf{AP$^{\text{mask}}$}} &
      {\textbf{AP$^{\text{box}}_{\mathrm{r}}$}} &
      {\textbf{AP$^{\text{mask}}_{\mathrm{r}}$}} \\
    \midrule
    (a) Real-only\hspace*{24mm} & 47.5 & 42.3 & 41.4 & 36.8 \\ \hdashline
    (b) $M_\text{teacher}$ only & 49.4 & 43.9 & \textbf{47.9} & 43.0 \\
    (c) $M_\mathcal{P}$ only & 49.9 & 44.9 & 45.3 & 41.7 \\
    (d) $M_\text{teacher}$ and $M_\mathcal{P}$ & \textbf{50.3} & \textbf{45.1} & 47.5 & \textbf{43.8} \\
    \bottomrule
  \end{tabular}}
  \caption{
\textbf{Ablation Study on T2I Pseudo-label Supervision.} 
We evaluate the impact of different pseudo-label sources: offline pseudo labeler ($M_\mathcal{P}$), EMA teacher ($M_\text{teacher}$), and their combination. 
The results show that $M_\mathcal{P}$ provides stable supervision for common categories, $M_\text{teacher}$ adaptively improves rare-class coverage, and combining both leverages complementary strengths.
  }
  \label{tab:ablation_t2i}
  \vspace{-6mm}
\end{table}

\subsection{Ablation Study}
\subsubsection{Effect of EMA-Teacher Labeling.} To evaluate the effectiveness of our EMA-teacher framework, we conducted an ablation study on T2I data labeling, as shown in Table~\ref{tab:ablation_t2i}. 
Specifically, we trained separate instance segmentation models using either the offline pseudo labels ($A_m^\mathcal{P}$) or the EMA teacher-generated labels ($\tilde{A}_m^\text{teacher}$) as supervision. 
The model trained with $A_m^\mathcal{P}$ achieves relatively higher AP on common categories, reflecting the stable supervision provided by the offline pseudo labels. 
In contrast, the model trained with $\tilde{A}_m^\text{teacher}$ leverages pseudo labels that are progressively updated by the EMA teacher throughout training, adaptively correcting errors and improving AP on underrepresented rare classes.
Combining both $M_{\mathcal{P}}$ and $M_\text{teacher}$ leverages their complementary strengths, maintaining high accuracy for common categories while improving rare-class coverage, as visually illustrated in Fig.~\ref{fig:T2I_merge}.

\begin{figure*}[t]
\begin{center}
\includegraphics[width=1.0\linewidth]{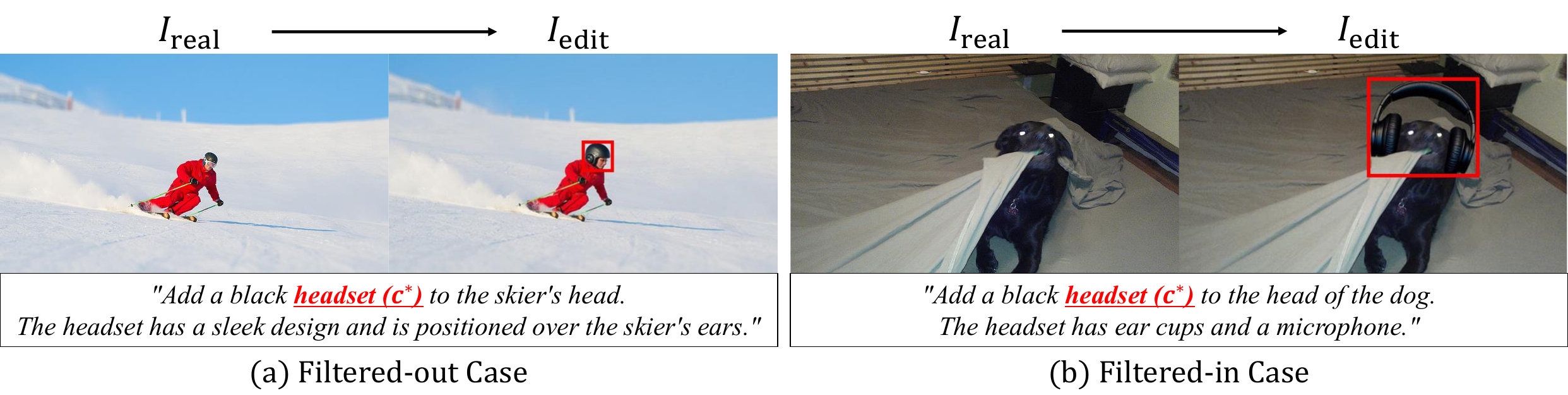}
\vskip -0.1in
\caption{
\textbf{Qualitative Examples of the VRAIN Verification Stage.}
(a) A `helmet' is added instead of the requested `headset' ($c^*$), creating a semantic mismatch, and is thus filtered out; (b) The target `headset' is correctly added, passing the verification.
}
\label{fig:vlm_filtering}
\end{center}
\vspace{-4mm}
\end{figure*}

\begin{table}[h]
  \centering
  \vspace{-2mm}
  \resizebox{0.65\columnwidth}{!}{%
  \begin{tabular}{p{3.5cm} c c c c}
    \toprule
    \textbf{Method} & \textbf{Precision} & \textbf{Recall} & \textbf{FP Rate} & \textbf{FN Rate} \\
    \midrule
    (a) CLIP (Crop) & 0.886 & 0.623 & 0.187 & 0.377 \\
    (b) VLM (Crop) & 0.975 & 0.521 & \textbf{0.031} & 0.478 \\
    (c) VLM (Ours) & \textbf{0.984} & \textbf{0.846} & 0.033 & \textbf{0.154} \\
    \bottomrule
  \end{tabular}}
  \caption{
  \textbf{VLM Verification Reliability on LVIS Validation.}
  Compared to prior CLIP-based verification, our spatially grounded VLM verifier achieves a superior precision–recall trade-off with lower FP and FN rates.
}
  \vspace{-6mm}
  \label{tab:rebuttal_vlm_verification}
\end{table}

\begin{table}[h!]
  \centering
  \resizebox{0.65\columnwidth}{!}{%
  \begin{tabular}{l c c c c}
    \toprule
    \textbf{Method} &
      {\textbf{AP$^{\text{box}}$}} &
      {\textbf{AP$^{\text{mask}}$}} &
      {\textbf{AP$^{\text{box}}_{\mathrm{r}}$}} &
      {\textbf{AP$^{\text{mask}}_{\mathrm{r}}$}} \\
    \midrule
    (a) Real-only\hspace*{24mm} & 47.5 & 42.3 & 41.4 & 36.8 \\ \hdashline
    (b) $\mathcal{D}_\text{I2I}$ only (w/o verification) & 47.4 & 42.5 & 39.5 & 36.0 \\
    (c) $\mathcal{D}_\text{I2I}$ only (VRAIN) & \textbf{48.1} & \textbf{43.3} & \textbf{42.9} & \textbf{39.5} \\
    \bottomrule
  \end{tabular}}
  \caption{
  \textbf{Ablation Study on VLM Verification in VRAIN.}
We compare I2I data with and without VLM-based filtering. Removing the verification step (b) reduces both overall and rare-class AP, indicating that the VLM filter is crucial for retaining semantically correct and relevant instances.
}
  \label{tab:ablation_i2i_vlm}
\vspace{-8mm}
\end{table}

\subsubsection{Reliability of VRAIN's Verification and Annotation.}
To quantitatively assess the reliability of our VLM-based verification module in VRAIN, we construct a controlled perturbation benchmark from the LVIS validation set (244k instances).
We generate negative samples by randomly perturbing instance class labels with 30\% probability while preserving the original images and bounding boxes.
We measure false positive (FP) rate when the verifier incorrectly accepts perturbed labels and false negative (FN) rate when it rejects correct ones.
We compare three verification strategies:
(a) CLIP similarity on crops (``a photo of [class]”),
(b) crop-based VLM verification (``Is the image a [class]?”),
and (c) our spatially grounded red-box VLM (``Is the red bounding box a [class]?”).
As shown in Table~\ref{tab:rebuttal_vlm_verification}, the VLM verifier significantly reduces the FP rate compared to CLIP (0.187 $\rightarrow$ 0.033).
Moreover, explicitly providing spatial grounding significantly reduces the FN rate (0.478 $\rightarrow$ 0.154) compared to the crop-based VLM, achieving a balanced trade-off between precision and recall.
We further evaluate the impact of the verification step by generating I2I data without applying VLM filtering.
As shown in Table~\ref{tab:ablation_i2i_vlm}-(b), generating I2I data without VLM verification degrades both overall and rare-class performance, highlighting the importance of semantic consistency filtering for reliable synthetic supervision.
Qualitative examples of filtered and retained samples are shown in Fig.~\ref{fig:vlm_filtering}.

Beyond semantic verification, we validate the mask annotation quality of VRAIN using the ORIDa benchmark~\cite{kim2025orida}, which contains 108k real before/after image pairs with physically placed objects and ground-truth annotations.
By treating the before and after images as $I_\text{real}$ and $I_\text{edit}$ respectively, we apply our annotation pipeline and compare the resulting masks against the ground-truth.
On this benchmark, our method achieves 0.925 mAP on mask annotations, demonstrating high annotation fidelity comparable to human-level quality~\cite{gupta2019lvis,tschirschwitz2025label}.

\subsubsection{Contribution of the I2I Data.}
Finally, we assessed the overall contribution of the generated high-fidelity I2I data ($\mathcal{D}_\text{I2I}$) to the hybrid training framework. 
As shown in Table~\ref{tab:ablation_i2i}, excluding $\mathcal{D}_\text{I2I}$ leads to a noticeable performance drop, particularly for rare categories (AP$_r^\text{box}$ 49.1 → 47.5, AP$_r^\text{mask}$ 44.0 → 43.8).
This confirms that $\mathcal{D}_\text{I2I}$ successfully provides targeted rare-class coverage that complements the T2I data.
By integrating these reliable, instance-level annotations, the model learns more robust representations for underrepresented classes, further boosting overall accuracy.

\begin{table}[t]
  \centering
  \resizebox{0.65\columnwidth}{!}{%
  \begin{tabular}{l c c c c}
    \toprule
    \textbf{Method} &
      {\textbf{AP$^{\text{box}}$}} &
      {\textbf{AP$^{\text{mask}}$}} &
      {\textbf{AP$^{\text{box}}_{\mathrm{r}}$}} &
      {\textbf{AP$^{\text{mask}}_{\mathrm{r}}$}} \\
    \midrule
    (a) Real-only\hspace*{24mm} & 47.5 & 42.3 & 41.4 & 36.8 \\ \hdashline
    (b) w/o $\mathcal{D}_\text{I2I}$ & 50.3 & 45.1 & 47.5 & 43.8 \\
    (c) Ours & \textbf{50.7} & \textbf{45.2} & \textbf{49.1} & \textbf{44.0} \\
    \bottomrule
  \end{tabular}}
  \caption{
  \textbf{Ablation Study on I2I Dataset.}
Excluding $\mathcal{D}_\text{I2I}$ lowers overall and rare-class performance, emphasizing the value of high-fidelity, rare-class-targeted I2I data that collaboratively complements T2I supervision.
}
  \label{tab:ablation_i2i}
\vspace{-7mm}
\end{table}

\section{Conclusion}
We present a unified data synthesis framework that bridges T2I generation and context-aware I2I editing for large-vocabulary instance segmentation. 
Our approach effectively couples the diversity and scalability of T2I synthesis with the realism and precise rare-class supervision of I2I editing. 
Through prompt-consistent filtering and teacher-student refinement, the T2I branch achieves reliable label quality despite domain gaps, while the proposed VRAIN framework provides high-fidelity, contextually coherent rare-class augmentation that reinforces the teacher-student learning process. 
Experiments on LVIS show that this hybrid strategy surpasses existing T2I and paste-based baselines, particularly on rare categories, and scales favorably with larger backbones. 
We believe this integration of complementary generative paradigms offers a promising approach to boosting model performance across diverse and underrepresented categories in long-tailed instance segmentation.

\subsubsection{Limitation.}
Our hybrid approach relies on off-the-shelf generative models, which do not perfectly adhere to input instructions. For example, the T2I branch uses prompt-consistent filtering to retain relevant instances, but it may still fail to realize certain categories specified in the prompt. 
Similarly, the I2I can ignore positional instructions and place objects incorrectly.
Although these approaches help mitigate label noise and contextual inconsistency, achieving perfect fidelity and precise instance-level control across all synthetic data remains a persistent challenge.

\input{sec/sup}

\clearpage  


%
%
\bibliographystyle{splncs04}
\bibliography{main}
\end{document}

%% file: sec/sup.tex
\renewcommand{\thetable}{\Alph{table}}
\setcounter{table}{0}
\renewcommand{\thefigure}{\Alph{figure}}
\setcounter{figure}{0}

\begin{center}
    \Large \textbf{Supplementary Material of \\ TMI: Text-to-Image Meets Image-to-Image for Complementary Data Synthesis to Boost Long-Tailed Instance Segmentation}
\end{center}
\vspace{10pt}

\appendix

\section{Additional Experiments}
\subsection{Ablation Study on Synthetic Data Scale and Composition}
We analyze the effect of the scale and composition of synthetic data through an ablation study (Table~\ref{tab:ablation_data}).
First, synthetic data consistently improves performance over the real-only baseline (a), confirming the benefit of augmenting real images with our synthetic samples (b--e).
Second, increasing the scale of synthetic data enhances performance: using the full T2I and I2I sets (e) outperforms the 50\% subset (d), demonstrating favorable scaling of the synthetic data.
Finally, combining the two synthetic branches yields the best model. The I2I branch contributes rare-focused target selection, high-quality instance-level labels, and context-aware realism, while the T2I branch, annotated via our combined offline and online adaptive labeling pipeline, provides broad category coverage and high-quality pseudo labels. Together, this combination boosts both overall and rare-class performance over each individual branch (b) and (c) alone.

\begin{table}[h!]
    \centering
    \resizebox{0.8\columnwidth}{!}{%
    \begin{tabular}{l | l l | c c c c}
        \toprule
        \makecell[l]{\textbf{Method}} &
        \makecell{\textbf{\# T2I }} &
        \makecell{\textbf{\# I2I }} &
        \makecell{\textbf{AP$^{\text{box}}$}} &
        \makecell{\textbf{AP$^{\text{mask}}$}} &
        \makecell{\textbf{AP$^{\text{box}}_{\mathrm{r}}$}} &
        \makecell{\textbf{AP$^{\text{mask}}_{\mathrm{r}}$}} \\
        \midrule
        \makecell[l]{(a) Real-only} & \makecell[l]{$0\%$} & \makecell[l]{$0\%$} & 47.5 & 42.3 & 41.4 & 36.8 \\ \hdashline
        \makecell[l]{(b) T2I-only} & \makecell[l]{$100\%$} & \makecell[l]{$0\%$} & 50.3 & 45.1 & 47.5 & 43.8 \\
        \makecell[l]{(c) I2I-only} & \makecell[l]{$0\%$} & \makecell[l]{$100\%$} & 48.1 & 43.3 & 42.9 & 39.5 \\
        \makecell[l]{(d) T2I + I2I (50\%)} & \makecell[l]{$50\%$} & \makecell[l]{$50\%$} & 50.2 & 45.0 & 47.5 & 43.1 \\
        \makecell[l]{(e) T2I + I2I (100\%; Ours)} & \makecell[l]{$100\%$} & \makecell[l]{$100\%$} & \textbf{50.7} & \textbf{45.2} & \textbf{49.1} & \textbf{44.0} \\
        \bottomrule
        \end{tabular}}
        \caption{
        \textbf{Ablation Study on Synthetic Data Scale and Composition.} 
        Synthetic data improves performance over the real-only baseline (a), and larger amounts yield further gains (d → e). Adding I2I to T2I-only (b → e) boosts rare-class AP, while adding T2I to I2I-only (c → e) benefits from our merged offline-online T2I supervision.
        }
    \label{tab:ablation_data}
\end{table}

\newpage

\subsection{Analysis on Progressive Teacher Adaptation}
We introduce an EMA-based teacher-student framework combined with text-consistent filtering to progressively refine the pseudo labels on T2I data.
To assess the effectiveness of our design, we focus on two key aspects: (1) progressive, adaptive pseudo labels generated by the EMA teacher, and (2) the use of prompt-inconsistent predictions as \textit{unlabeled} instances.

\subsubsection{\textit{Does the teacher effectively adapt to the T2I domain, and is it important?}}
In our training strategy for the T2I dataset, the teacher model $M_\text{teacher}$ is updated via EMA and gradually adapts to the T2I domain while being guided by rare-class-focused I2I supervision, producing online pseudo labels $\tilde{A}_m^\text{teacher}$.
To verify the effectiveness of this progressive adaptation, we compare it against a control variant where the teacher's pseudo labels $\tilde{A}_m^\text{teacher}$ on T2I data are \emph{frozen} at initialization.
All other architecture and training hyperparameters are kept identical; the only difference is whether T2I pseudo labels are progressive or static.
As shown in Table~\ref{tab:experiment_frozen_teacher}, the fixed-label variant consistently underperforms our progressive teacher across all AP metrics, with particularly pronounced degradation on rare categories. 
This demonstrates that progressively adapting the EMA teacher to the T2I domain and refining pseudo labels over training is crucial for improving performance, particularly on underrepresented categories.

\begin{table}[h!]
  \centering
  \resizebox{0.65\columnwidth}{!}{%
  \begin{tabular}{l c c c c}
    \toprule
    \textbf{Method} &
      {\textbf{AP$^{\text{box}}$}} &
      {\textbf{AP$^{\text{mask}}$}} &
      {\textbf{AP$^{\text{box}}_{\mathrm{r}}$}} &
      {\textbf{AP$^{\text{mask}}_{\mathrm{r}}$}} \\
    \midrule
    (a) Real-only\hspace*{24mm} & 47.5 & 42.3 & 41.4 & 36.8 \\ \hdashline
    (b) $\tilde{A}_m^\text{teacher}$ (\emph{frozen}) & 48.8 & 43.8 & 45.5 & 41.2 \\
    (c) $\tilde{A}_m^\text{teacher}$ (\emph{progressive}) & \textbf{49.4} & \textbf{43.9} & \textbf{47.9} & \textbf{43.0} \\
    \bottomrule
  \end{tabular}}
  \caption{
  \textbf{Experiments on the Effectiveness of Adaptive Pseudo Label in T2I Dataset.}
We compare static T2I pseudo labels frozen at initialization (fixed-label) versus refreshed pseudo labels from the EMA-updated teacher (progressive). Refreshing pseudo labels over training yields consistent AP gains, especially on rare categories.
}
  \label{tab:experiment_frozen_teacher}
\vskip -0.2in
\end{table}

\subsubsection{\textit{Is it beneficial to treat high-confident prompt-inconsistent predictions as unlabeled?}}
We analyze the role of unlabeled instances in our T2I training setup.
Recall that we sample a category set $C_m$ from the long-tailed taxonomy to construct the text prompt, but the synthesized T2I images often include additional instances that are not explicitly specified in the prompt.
Our method applies text-consistent filtering by utilizing the set $C_m$, keeping only prompt-consistent instances whose predicted class belongs to $C_m$.
While this improves the precision of T2I pseudo labels, it also removes many visually present objects outside $C_m$, leading to a sparsely annotated object detection (SAOD)~\cite{niitani2019sampling, wang2021co} scenario where frequently occurring objects can remain unannotated and are treated as background during training.

To alleviate this issue, we reuse predictions that are high-confidence yet inconsistent with the prompt as unlabeled instances.
These instances are excluded from the classification loss but included in the box and mask losses~\cite{zhou2021probabilistic_ceternet2}, providing pure localization supervision that encourages the model to recognize that ``there is an object here'' without enforcing a potentially incorrect class label.
To verify the effectiveness of this approach, we compare two variants against our method on T2I data, as reported in Table~\ref{tab:experiment_unlabel}: (b) \textit{no-unlabeled} variant that discards all prompt-inconsistent predictions and relies only on prompt-consistent labels, (c) \textit{label-as-is} variant that treats high-score but prompt-inconsistent predictions as labeled instances with the teacher's predicted class, and (d) our method, which instead treats high-score but prompt-inconsistent predictions as unlabeled instances.
The \textit{no-unlabeled} variant (b) suffers a noticeable drop in overall AP, with larger gaps on categories that frequently appear but are not prompted, highlighting that simply ignoring prompt-inconsistent predictions is ineffective under a sparsely annotated regime.
The \textit{label-as-is} variant, while providing additional supervision, underperforms our method and even falls below the \textit{no-unlabeled} variant, indicating that directly treating prompt-inconsistent predictions as labeled instances is sensitive to label noise from incorrect class assignments, underscoring the necessity of prompt-consistent filtering.
In contrast, our unlabeled instance strategy achieves the best performance among the three, demonstrating that combining the prompt-consistent filtering for precise T2I labels with the use of high-confidence prompt-inconsistent predictions solely for localization supervision is a robust and effective choice for handling the sparsely annotated nature of T2I data.

\begin{table}[h!]
  \centering
  \resizebox{0.65\columnwidth}{!}{%
  \begin{tabular}{l c c c c}
    \toprule
    \textbf{Method} &
      {\textbf{AP$^{\text{box}}$}} &
      {\textbf{AP$^{\text{mask}}$}} &
      {\textbf{AP$^{\text{box}}_{\mathrm{r}}$}} &
      {\textbf{AP$^{\text{mask}}_{\mathrm{r}}$}} \\
    \midrule
    (a) Real-only\hspace*{24mm} & 47.5 & 42.3 & 41.4 & 36.8 \\ \hdashline
    (b) $M_\text{teacher}$ only (w/o unlabel) & 47.3 & 42.4 & 44.8 & 41.1 \\
    (c) $M_\text{teacher}$ only (unlabel $\rightarrow$ label) & 46.8 & 41.8 & 42.2 & 39.1 \\
    (d) $M_\text{teacher}$ only & \textbf{49.4} & \textbf{43.9} & \textbf{47.9} & \textbf{43.0} \\
    \bottomrule
  \end{tabular}}
  \caption{
  \textbf{Experiments on the Effectiveness of Unlabeled Pseudo Labels.}
Treating high-confidence prompt-inconsistent predictions as unlabeled (d) improves both overall and rare-class AP. Discarding them (b) or treating them as labeled instances (c) leads to lower performance, highlighting the importance of unlabeled supervision in sparsely annotated T2I data.
}
  \label{tab:experiment_unlabel}
\end{table}

\newpage

\section{Implementation Details}\label{implementation_details}
\subsection{T2I Generation}
\subsubsection{Textual Prompt Set Generation.}
To generate the textual prompt set $\mathcal{T}$ of size $N_\text{text}$, we create each text prompt by randomly selecting $l$ categories from the total category set $C_\text{all}$, where $l$ is chosen independently for each prompt in the range 5--10 to ensure diverse compositions. 
These selected classes form the set $C_m$ for that prompt, and are provided to GPT-4o~\cite{hurst2024gpt4o} using the following instruction template:

\begin{myverb}
Create a natural, concise image generation prompt using the selected object classes.
The prompt should describe a realistic scene where these objects naturally coexist.

Selected classes: {(*@\textbf{selected\_classes}@*)}

CRITICAL REQUIREMENTS:
1. MANDATORY: Include EVERY SINGLE selected object class in the prompt - NO EXCEPTIONS
2. Each object must be explicitly mentioned by name in the final prompt
3. Create a natural scene description (not just object listing)
4. Keep total prompt under 40 words
5. Don't specify colors - let AI choose naturally
6. Focus on realistic spatial relationships and context
7. End with simple lighting/photography style

VERIFICATION CHECKLIST - Before submitting, ensure:
- Every object from the selected classes list appears in the prompt
- No object is missing or omitted
- All objects are mentioned by their exact names

STRUCTURE:
- Describe a realistic scene where these objects belong together
- Mention spatial relationships (on, near, beside, etc.)
- Brief setting and atmosphere
- ENSURE ALL OBJECTS ARE INCLUDED

Example formats:
"A kitchen scene with ceramic bowl on wooden cutting board beside leather handbag and steel refrigerator, warm natural lighting, professional photography"

"Living room interior featuring armchair near coffee table with lamp and books, cozy atmosphere, soft lighting"

"Garden setting with watering can beside flower pot and gardening tools on wooden bench, natural daylight"

IMPORTANT: If you cannot fit all objects naturally in 40 words, prioritize including ALL objects over perfect grammar.

Please provide your response in the following JSON format:
```json
{
    "prompt": "your scene description (max 40 words, MUST include ALL objects)"
}
\end{myverb}

\begin{figure}[b!]
\vskip -0.8in
\begin{center}
\includegraphics[width=0.81\linewidth]{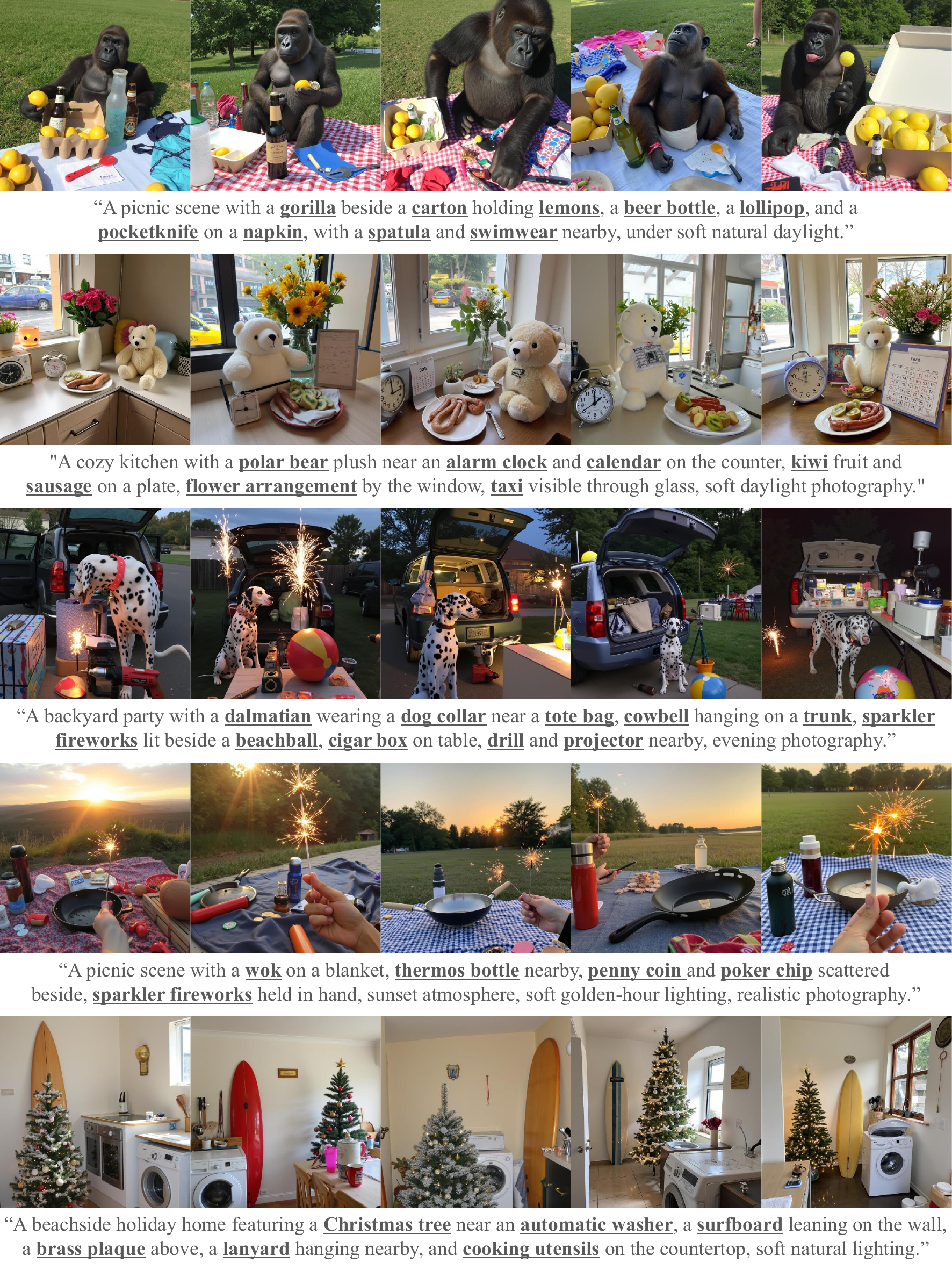}
\vskip -0.1in
\caption{\textbf{Examples of T2I Images.}
In each T2I prompt, boldfaced and underlined words indicate the target categories for generation.}
\label{fig:t2i_images}
\end{center}
\vskip -0.2in
\end{figure}

As an illustrative example, the randomly selected classes for a prompt may be \textit{gorilla}, \textit{lemon}, \textit{beer bottle}, \textit{pocketknife}, \textit{carton}, \textit{napkin/table napkin/serviette}, \textit{spatula}, \textit{lollipop}, and \textit{swimsuit/swimwear/bathing suit/swimming costume/bathing costume/swimming trunks/bathing trunks}.
Given this $\{$\texttt{selected\_classes}$\}$, GPT-4o generates a textual prompt such as ``A picnic scene with a \textit{gorilla} beside a \textit{carton} holding \textit{lemons}, a \textit{beer bottle}, a \textit{lollipop}, and a \textit{pocketknife} on a \textit{napkin}, with a \textit{spatula} and \textit{swimwear} nearby, under soft natural daylight'', which is a natural scene description that includes all selected object classes.
The prompt is then used by the T2I model $\Phi_{\text{T2I}}$~\cite{flux2024} to generate the corresponding image $I_m$. 
In practice, we generate five images per textual prompt to increase visual diversity.
Fig.~\ref{fig:t2i_images} shows representative examples of our T2I images created from the textual prompt set.

\newpage
\subsection{I2I Editing}
\subsubsection{Rare-class Proposal.}
In the rare-class proposal stage, we identify semantically suitable categories and determine their placement within a given real image. 
For each image $I_\text{real}$, we sample a candidate set $C_q$ of $Q$ classes from target rare categories $C_\text{I2I}$ using a softmax-based sampler weighted by previous generation counts. 
Specifically, the sampling weight $w(c)$ for each category $c \in C_\text{I2I}$ is computed as $w(c) = \exp(-\text{count}(c))$, where $\text{count}(c)$ is the total generation count for $c$.
The probability of selecting $c$ is $P(c) = w(c) / \sum_{c' \in C_\text{I2I}} w(c')$, and we draw $Q$ classes according to $P(c)$ to form the candidate set $C_q$. 
This procedure encourages balanced augmentation across rare classes.

Once $C_q$ is obtained, a Vision-Language Model (VLM)~\cite{zhu2025internvl3} is prompted using a natural language template to select the most semantically appropriate category $c^*$ and generate a corresponding instruction $inst^*$. 
The prompt template is as follows:
\begin{myverb}
I want to inpaint a semantically natural object into the provided image.
Please choose which of the following {(*@\textbf{Q}@*)} classes is most appropriate for pasting.
It is important to exclude any objects already present in the image.

Additionally, please provide an instruction in the format 
`Add <attribute_phrase> <class_name> to <location>. <extra_sentences>' 
to inpaint the selected class into the provided image.

Candidates: {(*@\textbf{class\_name\_str}@*)}

Answer using the following format without any explanation:

Objects: List the names of the objects in the provided image.
Appropriate class: <number>. <class_name>
Instruction: an instruction in the format 
    `Add <attribute_phrase> <class_name> to <location>. <extra_sentences>'

- <class_name>: the selected class.
- <attribute_phrase>: an attribute phrase (color, shape, brand, etc.).
- <location>: the location where the object should be added.
- <extra_sentences>: 1--2 sentences describing additional characteristics 
  of the object (e.g., color, material, distinguishing features).

Examples:
"Add a red chair to the left side of the room. 
The chair has wooden legs and a cushioned seat."

"Add a blue backpack to the floor. 
The backpack has multiple compartments and padded straps."

"Add a yellow folded umbrella near the child on the right side of the image. 
The umbrella has a curved handle and a cartoon character print."
\end{myverb}
Here, we set $Q=5$ candidate classes per image, and $\{$\texttt{class\_name\_str}$\}$ enumerates these classes with numbered labels for clarity. 
For instance, for the real image $I_\text{real}$ in the first row of Fig.~\ref{fig:i2i_ex1}, where a man is skiing downhill, suppose the candidate set $C_q$ contains the following five classes: \textit{garbage}, \textit{gondola/gondola boat}, \textit{comic book}, \textit{sparkler/sparkler fireworks}, and \textit{spear/lance}.
Then, $\{$\texttt{class\_name\_str}$\}$ would list them as:
\begin{enumerate}
    \item garbage
    \item gondola/gondola boat
    \item comic book
    \item sparkler/sparkler fireworks
    \item spear/lance
\end{enumerate}
From this set, the VLM selects the most semantically appropriate class $c^*$ and generates a corresponding natural language instruction $inst^*$, which is then used by the instruction-based editor to inpaint the object into the real image. 
For example, the VLM may choose \textit{comic book} as $c^*$ and produce the instruction ``Add a colorful comic book to the skier's left hand. The comic book has vibrant illustrations and text.'', which guides the editor to accurately place and render the object in the scene.
Following this instruction, the instruction-based editor produces the final edited image $I_\text{edit}$, showing the skier holding the inpainted comic book with coherent appearance and geometry.

\subsubsection{Verification.}
In the verification stage, we aim to ensure that each edited object is both semantically correct and visually accurate. 
Candidate detections from the edited image are first localized using a structural similarity (SSIM~\cite{ssim}) difference map or an open-vocabulary detector~\cite{fu2025llmdet}, producing a set of potential objects. 
To semantically verify these candidates, we employ a VLM~\cite{zhu2025internvl3} that is prompted with a visualization of the edited image where the candidate bounding box is overlaid in red (see Fig.~6 in the main text for visualization examples). 
The VLM is asked whether the object within the red box corresponds to the intended category using the following template:
\begin{myverb}
Is the red bounding box in the image a {(*@\textbf{class\_name}@*)}?
Answer in the following format without further explanation:
"Answer": Answer here with "Yes" if the red bounding box is a {(*@\textbf{class\_name}@*)}, "No" otherwise.
\end{myverb}
Here, $\{$\texttt{class\_name}$\}$ is $c^*$. Only candidates that receive a ``Yes'' from the VLM are retained, and instance masks are then obtained for these verified objects using SAM. 
This ensures that only semantically consistent objects are added to the final annotation.

\newpage

\section{Computational Costs}\label{computational_costs}
We analyze the computational cost of our data generation pipeline on a single NVIDIA H100 GPU. 
T2I generation takes 9.2s per sample, including 8s for image synthesis and 1.2s for annotation. 
I2I generation requires 20.6s per sample, consisting of rare-class proposal (1.96s), image editing (18s), and annotation with verification (0.64s). 
Notably, the verification overhead accounts for less than 3.5\% of the total I2I time, indicating that the proposed ``place-and-verify'' design does not introduce a significant computational bottleneck. 
Since this process is performed once offline for dataset construction, it remains substantially more scalable than manual annotation.

\section{Failure Modes and Statistics in VRAIN}\label{failcase}
\begin{figure}[h]
\begin{center}
\scriptsize
\vspace{-3mm}
\includegraphics[width=1.0\linewidth]{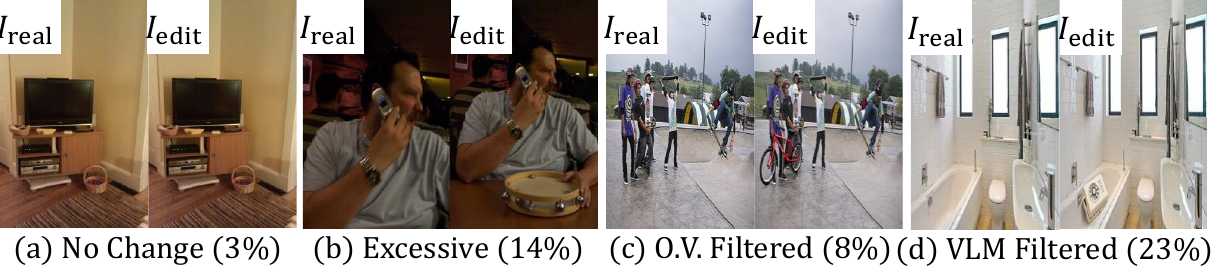}
\vspace{-4mm}
\caption{\textbf{VRAIN Filtering Visualization and Statistics.}
Representative failure cases and the rejection rate of each verification filter stage.
The instructions are: 
(a) ``\textit{Add a silver file (file tool) to the shelf below the television. The file has a metallic finish and is commonly used for smoothing surfaces.}''; 
(b) ``\textit{Add a wooden tambourine to the table in front of the man. The tambourine has a circular shape with jingle bells inside and a light brown color.}'' (the table is a hallucinated target location); 
(c) ``\textit{Add a red unicycle to the left side of the ramp. The unicycle has a black seat and silver spokes.}''; 
(d) ``\textit{Add a white phonebook to the left side of the sink. The phonebook has a rectangular shape and is similar in color to the bathroom tiles.}''
}
\label{fig:rebuttal_vrain_stats}
\end{center}
\vspace{-10mm}
\end{figure}

We visualize representative failure modes and filtering statistics in Fig.~\ref{fig:rebuttal_vrain_stats}. 
The rejected samples are categorized into four distinct types. 
Structural anomalies are filtered by SSIM: (a) no visible change where the editor fails to alter the image, and (b) excessive edits that disrupt the original scene geometry.
Object class errors are subsequently filtered by two-tage semantic verification process: (c) instances that fail to be localized by the open-vocabulary detector searching for the target class, and (d) fine-grained semantic mismatches further rejected by the VLM (\eg, synthesizing a phone instead of a phonebook). 
Fig.~\ref{fig:rebuttal_vrain_stats}-(b) illustrates an infeasible proposal where the VLM suggested ``\textit{Add a tambourine to the table}'' in a scene without a visible table. This led the editor to hallucinate a table, which was subsequently rejected by the SSIM-based structural consistency check. 
Although the overall acceptance rate is 52\%, this strict filtering strategy prioritizes quality over quantity, ensuring high-fidelity and context-consistent synthetic instances. 
Such a quality-first design is critical for bootstrapping T2I training, enabling more reliable learning of rare categories.

\newpage

\section{More Results}\label{more_results}
\subsection{T2I Dataset Analysis}
\begin{figure}[t]
\begin{center}
\includegraphics[width=0.9\linewidth]{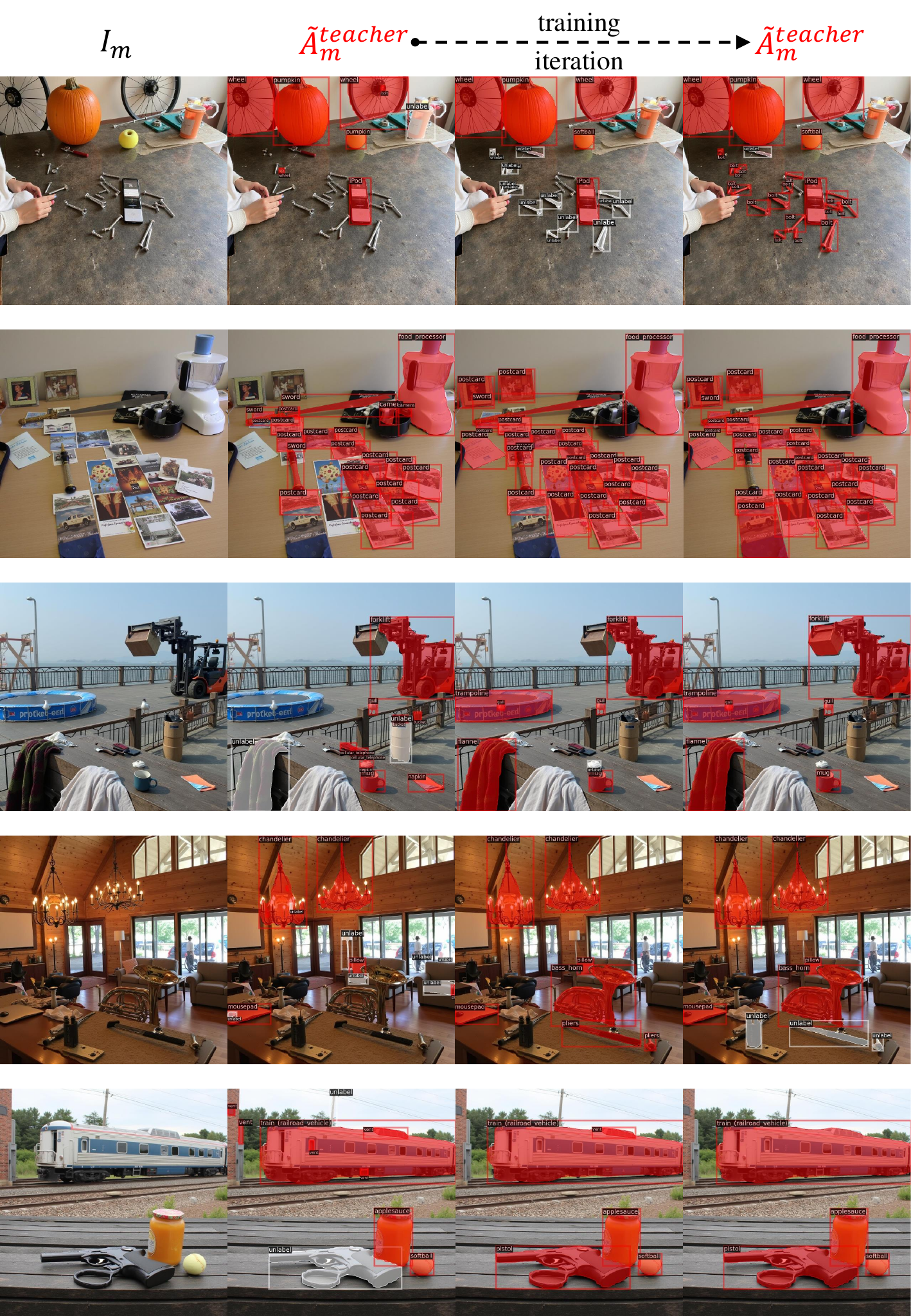}
\caption{\textbf{Examples of Pseudo-label Adaptation in T2I images.}
}
\label{fig:t2i_adapt_ex1}
\end{center}
\vskip -0.2in
\end{figure}

\begin{figure}[t]
\begin{center}
\includegraphics[width=0.9\linewidth]{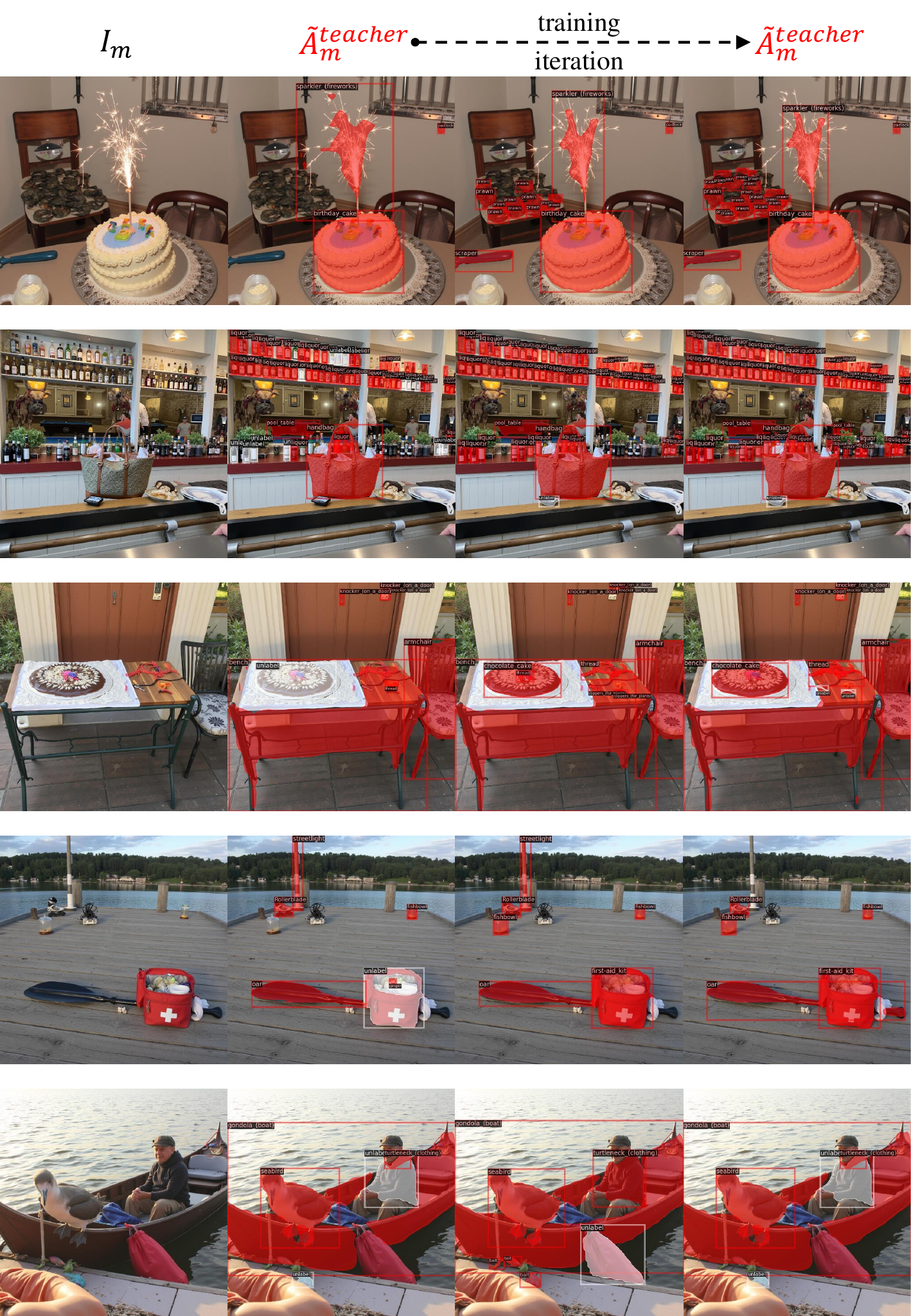}
\caption{\textbf{Examples of Pseudo-label Adaptation in T2I images.}
}
\label{fig:t2i_adapt_ex2}
\end{center}
\vskip -0.2in
\end{figure}

We provide more examples illustrating the evolution of teacher-generated pseudo labels on T2I images in Fig.~\ref{fig:t2i_adapt_ex1} and~\ref{fig:t2i_adapt_ex2}.
The first column shows the T2I image, and the following three columns depict the teacher’s predictions at early, intermediate, and late training stages, respectively, revealing the progressive refinement achieved via our EMA-based adaptation.

\begin{figure}[t]
\begin{center}
\includegraphics[width=1.0\linewidth]{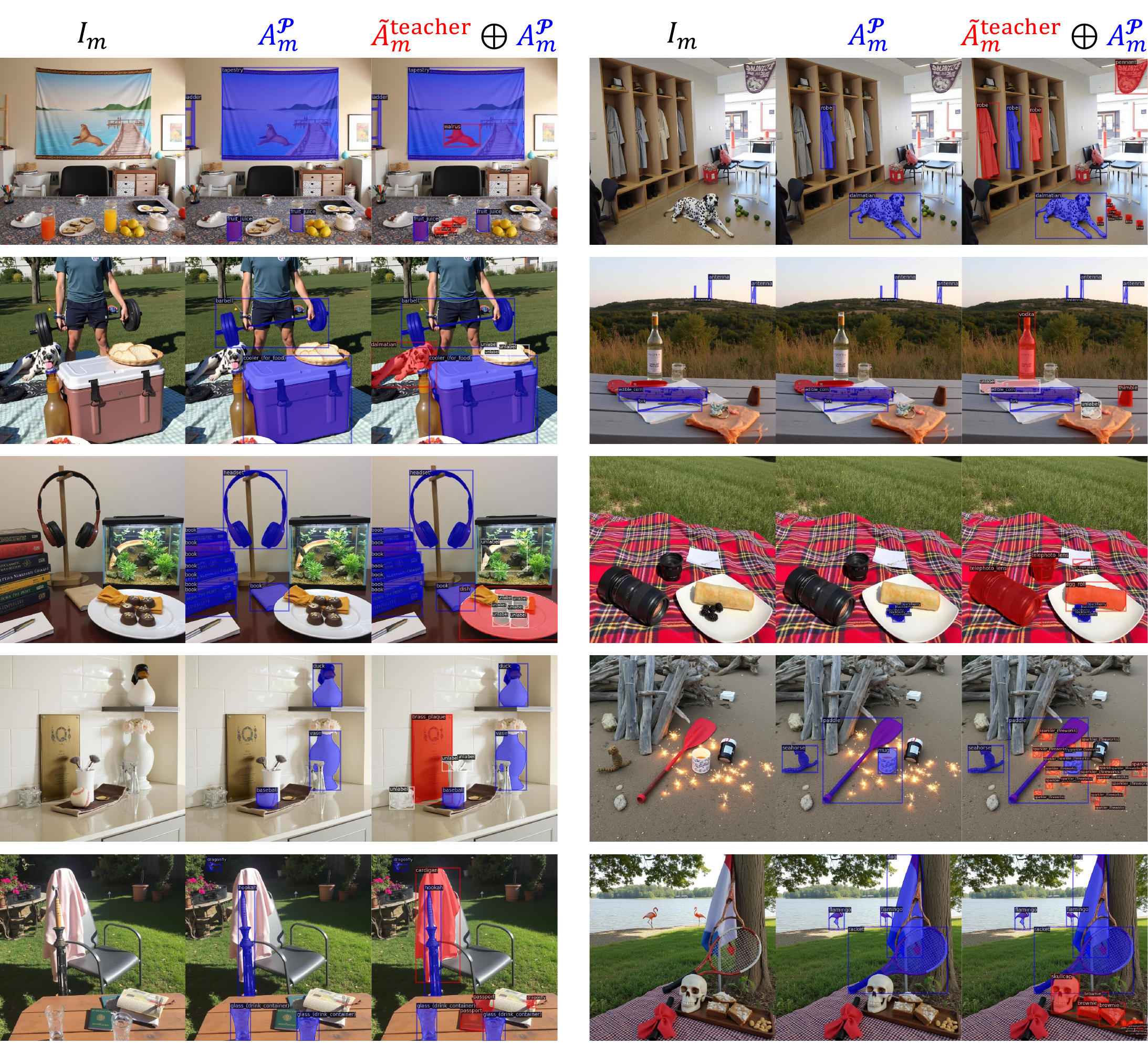}
\caption{\textbf{Examples of Final Pseudo-labels for T2I images.}
Teacher-generated pseudo-labels (\textcolor{red}{red}) are merged ($\oplus$) with offline pseudo-labels (\textcolor{blue}{blue}) to form the final supervision.
}
\label{fig:t2i_merge_ex1}
\end{center}
\vskip -0.2in
\end{figure}

\begin{figure}[t]
\begin{center}
\includegraphics[width=1.0\linewidth]{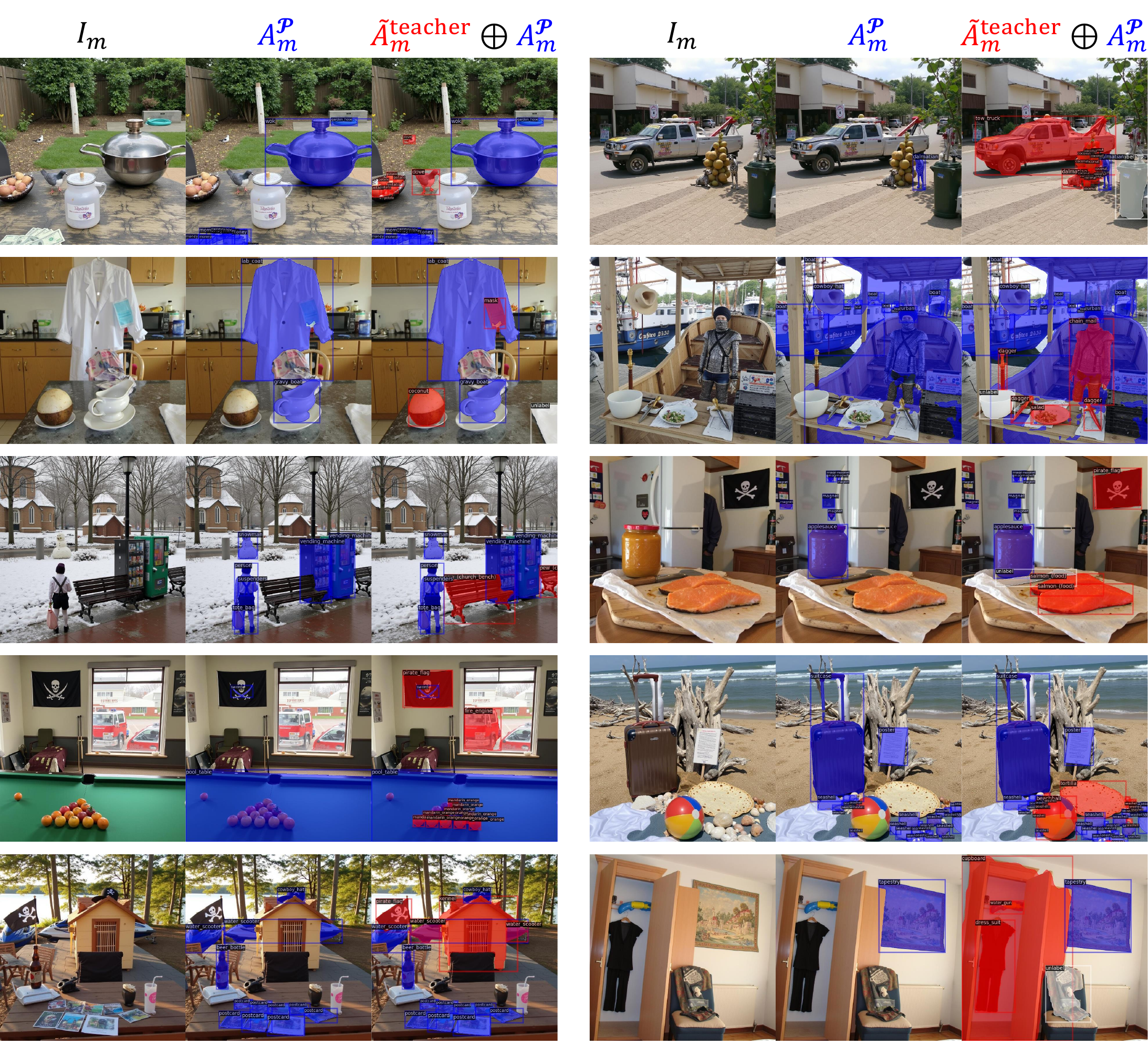}
\caption{\textbf{Examples of Final Pseudo-labels for T2I images.}
Teacher-generated pseudo-labels (\textcolor{red}{red}) are merged ($\oplus$) with offline pseudo-labels (\textcolor{blue}{blue}) to form the final supervision.
}
\label{fig:t2i_merge_ex2}
\end{center}
\vskip -0.2in
\end{figure}

We also present the constructed final supervision for T2I data in Fig.~\ref{fig:t2i_merge_ex1} and~\ref{fig:t2i_merge_ex2}.
The 1st and 4th columns display the T2I images, the 2nd and 5th columns show their offline pseudo labels, and the 3rd and 6th columns visualize the final merged labels, which combine the offline labels with the teacher’s refined predictions.
These examples highlight how both sources jointly contribute to producing reliable supervision for training on T2I images.

\subsection{I2I Dataset Examples}
\begin{figure}[t]
\begin{center}
\includegraphics[width=0.9\linewidth]{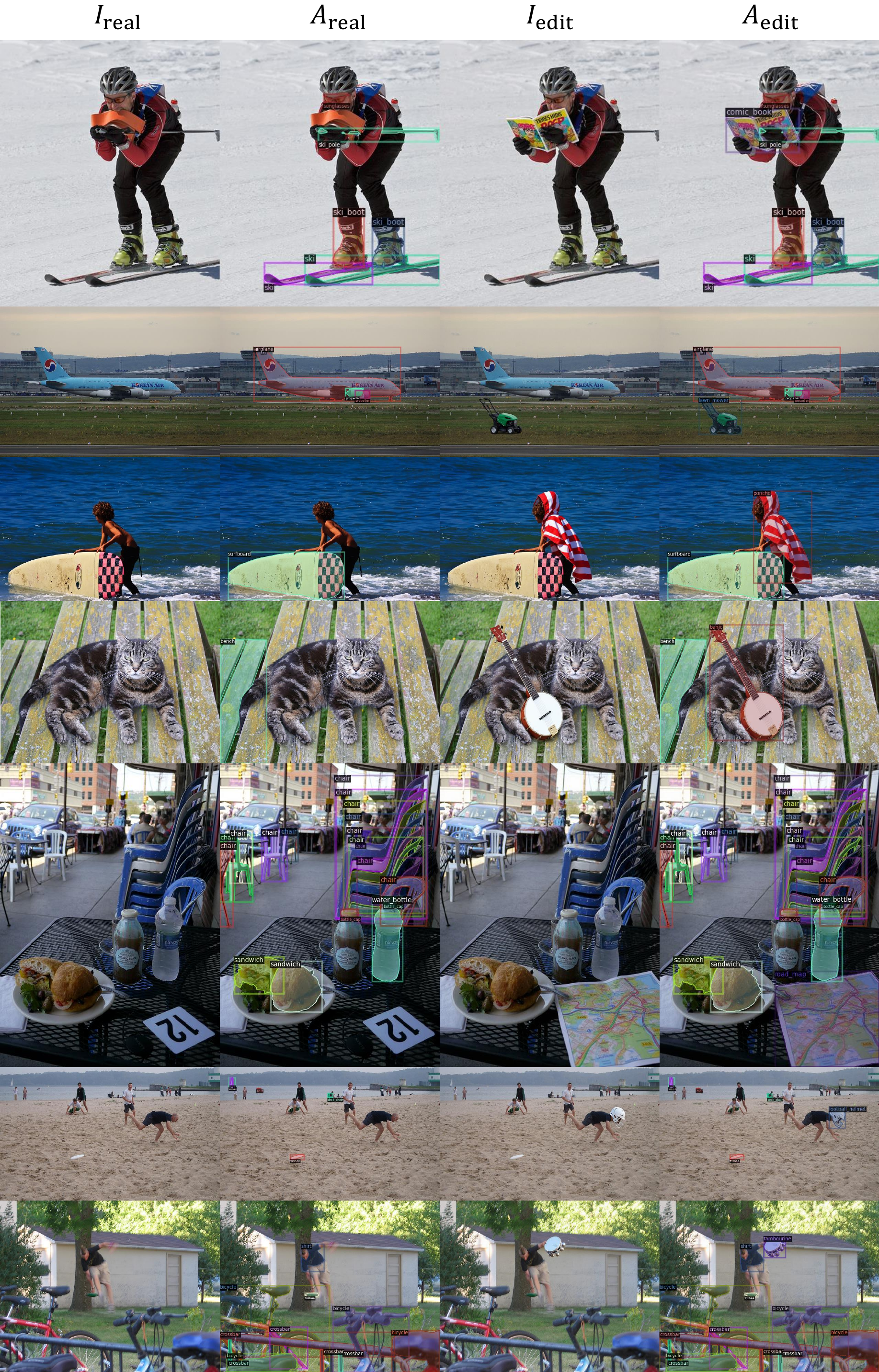}
\caption{\textbf{Examples of I2I Dataset.}
}
\label{fig:i2i_ex1}
\end{center}
\vskip -0.2in
\end{figure}

\begin{figure}[t]
\begin{center}
\includegraphics[width=0.9\linewidth]{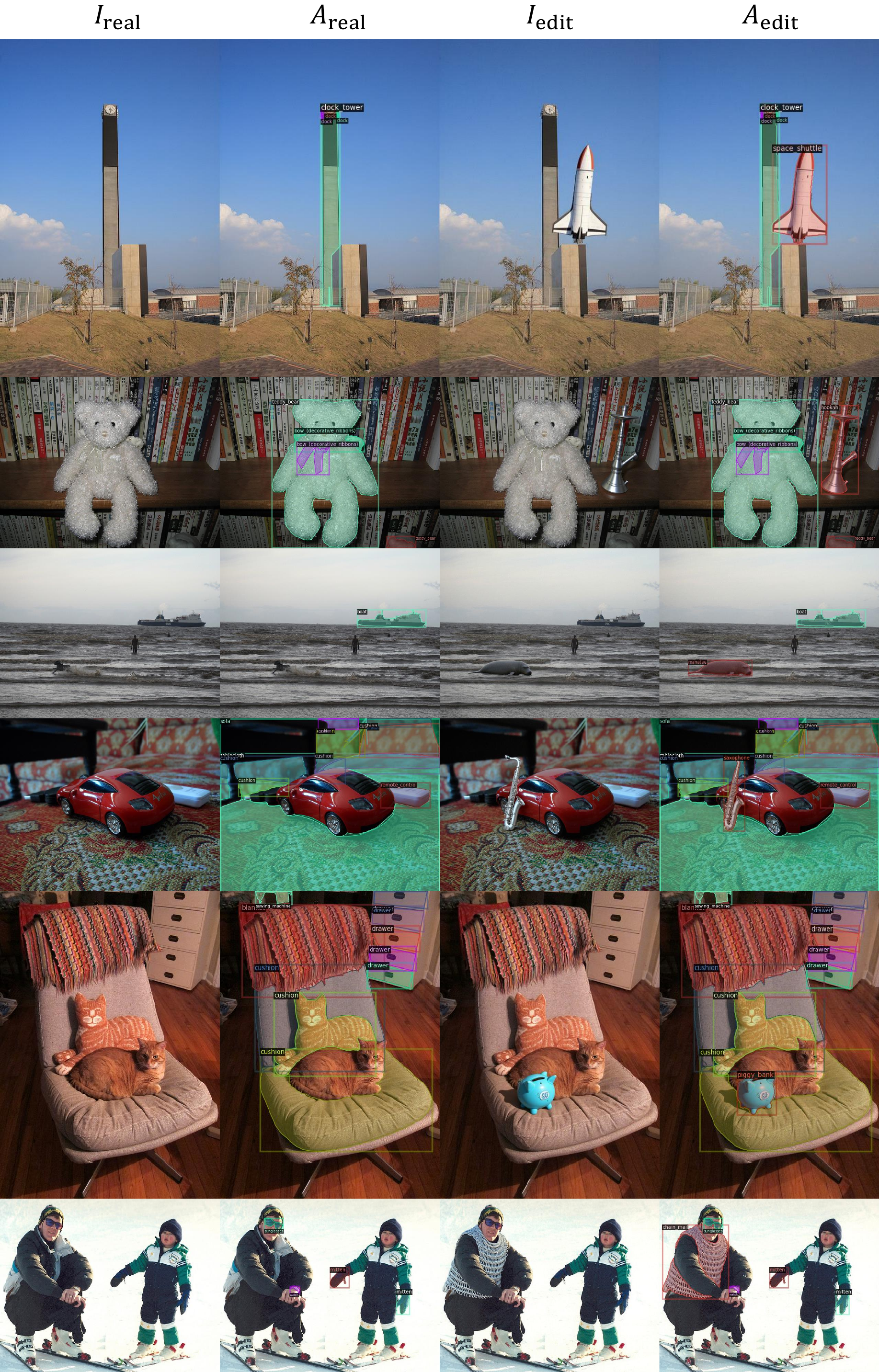}
\caption{\textbf{Examples of I2I Dataset.}
}
\label{fig:i2i_ex2}
\end{center}
\vskip -0.2in
\end{figure}

\begin{figure}[t]
\begin{center}
\includegraphics[width=0.9\linewidth]{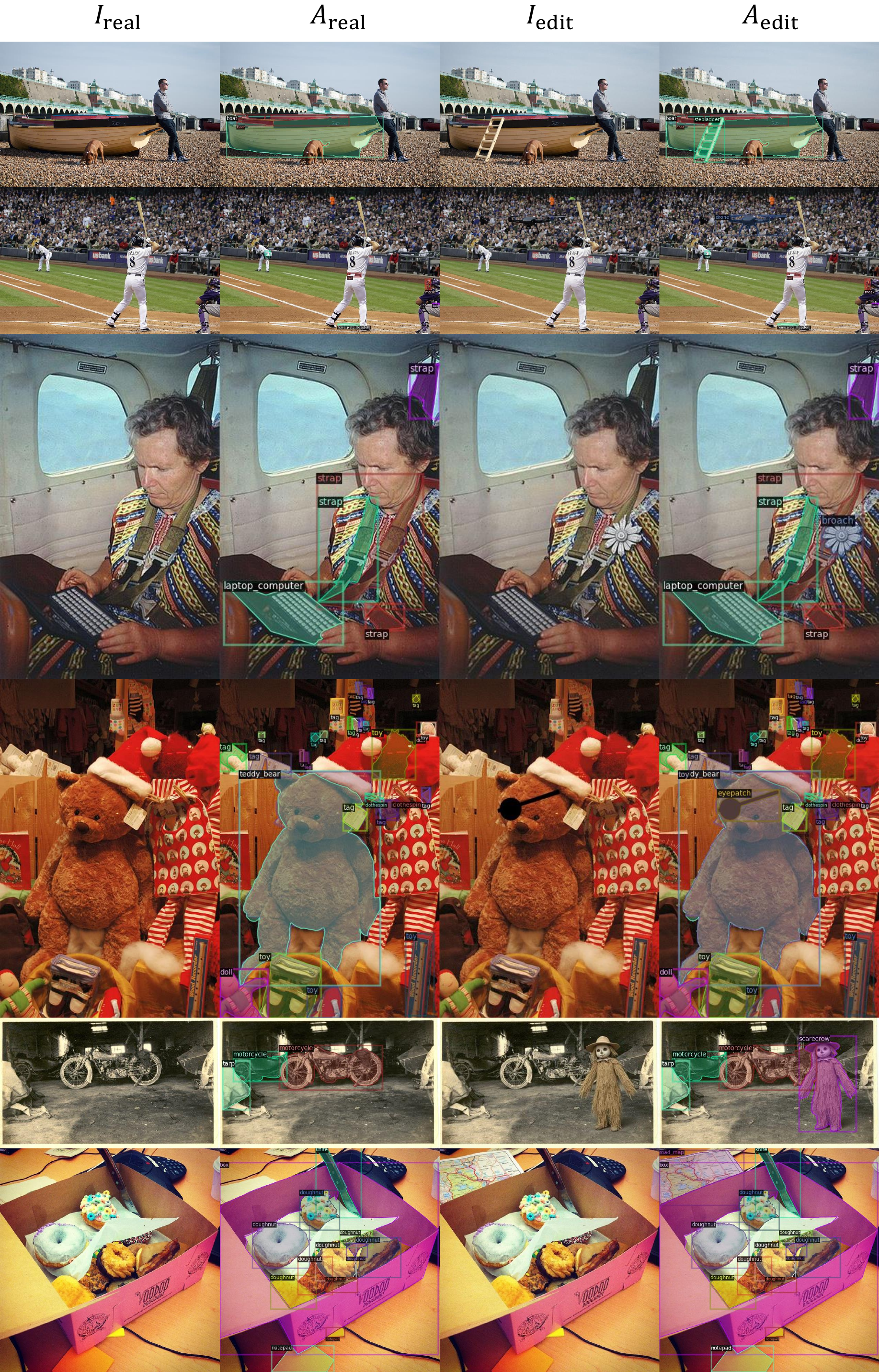}
\caption{\textbf{Examples of I2I Dataset.}
}
\label{fig:i2i_ex3}
\end{center}
\vskip -0.2in
\end{figure}

\begin{figure}[t]
\begin{center}
\includegraphics[width=0.9\linewidth]{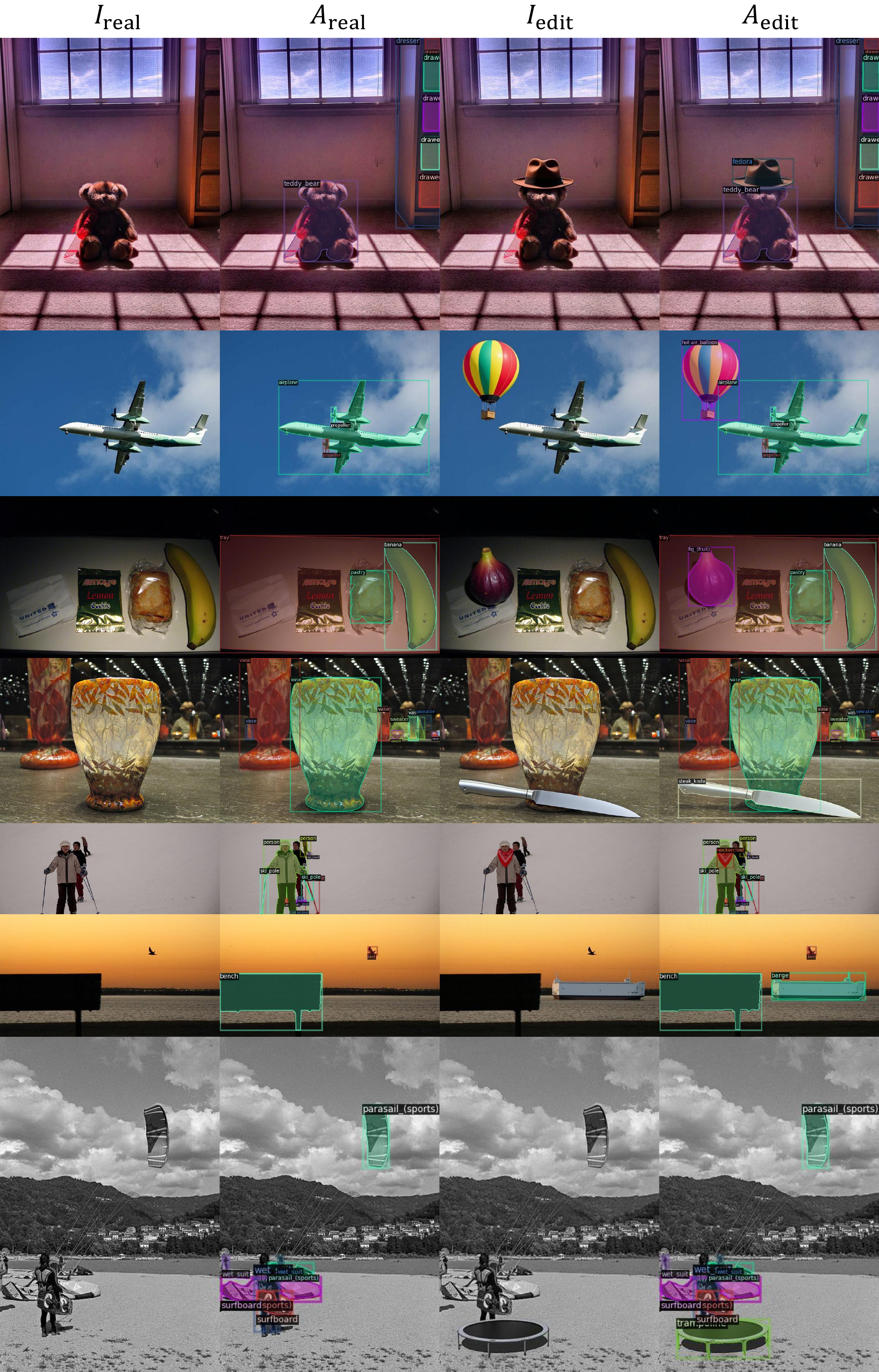}
\caption{\textbf{Examples of I2I Dataset.}
}
\label{fig:i2i_ex4}
\end{center}
\vskip -0.2in
\end{figure}

\begin{figure}[t]
\begin{center}
\includegraphics[width=0.9\linewidth]{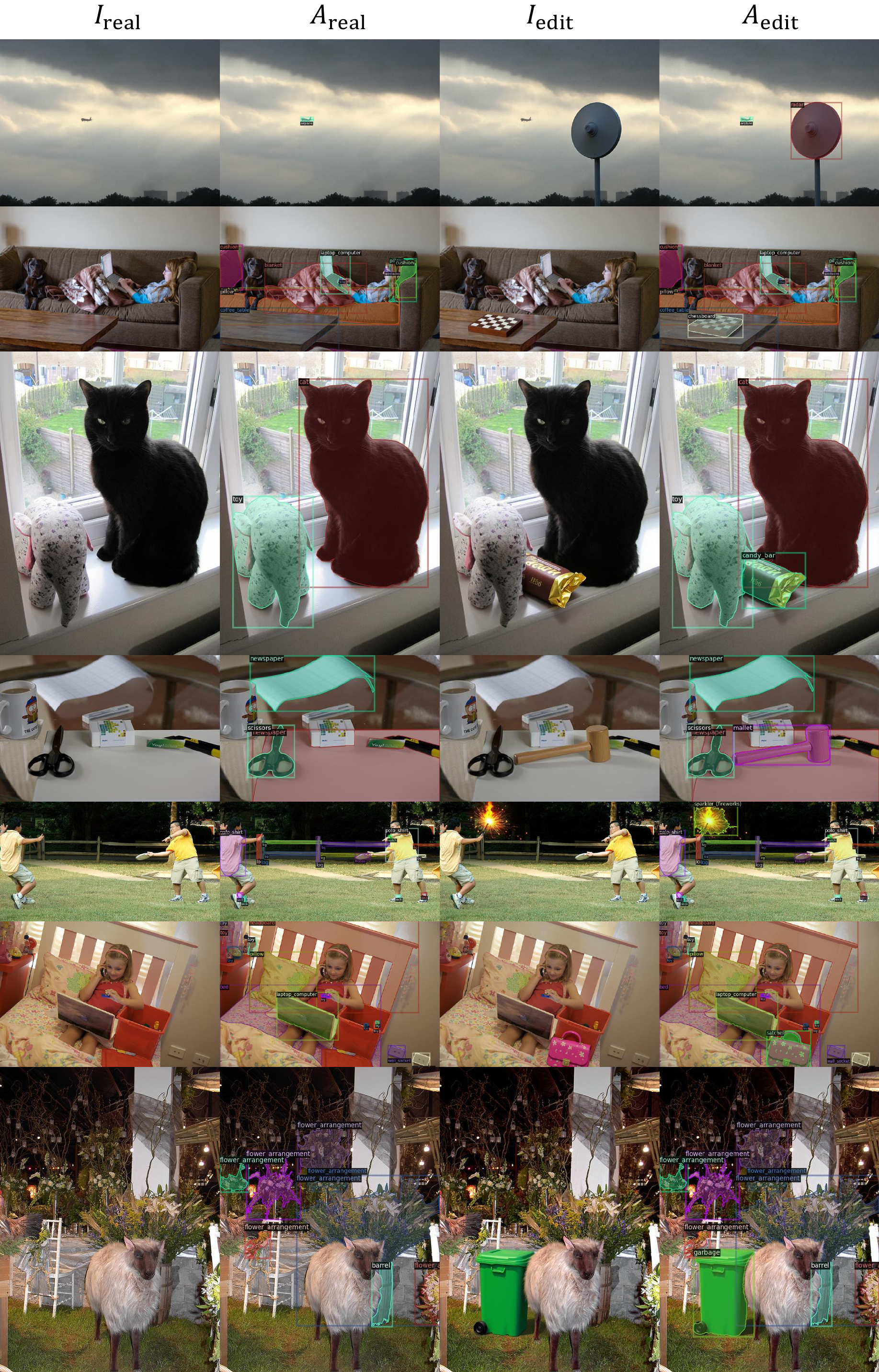}
\caption{\textbf{Examples of I2I Dataset.}
}
\label{fig:i2i_ex5}
\end{center}
\vskip -0.2in
\end{figure}

\begin{figure}[t]
\begin{center}
\includegraphics[width=0.9\linewidth]{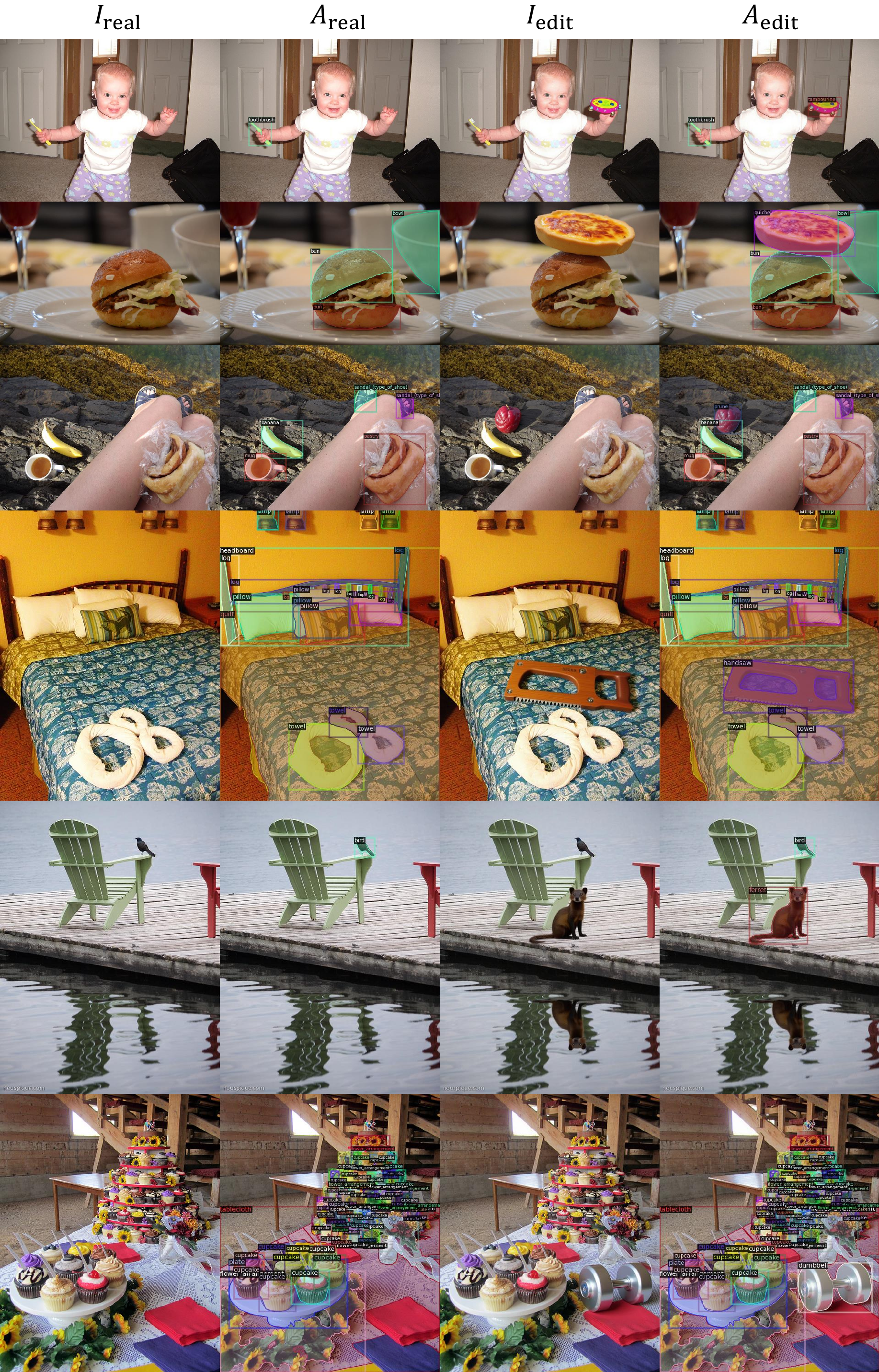}
\caption{\textbf{Examples of I2I Dataset.}
}
\label{fig:i2i_ex6}
\end{center}
\vskip -0.2in
\end{figure}

We present additional examples of data generated by our I2I editing pipeline. 
Each row consists of four columns: the original image $I_\text{real}$, the original annotation $A_\text{real}$, the edited image $I_\text{edit}$ (produced by the instruction-based editor $\Phi_{\text{I2I}}$), and the final edited annotation $A_\text{edit}$ (obtained after VLM verification and SAM mask integration). 
Examples across multiple scenes are shown in Fig.~\ref{fig:i2i_ex1}--\ref{fig:i2i_ex6}, illustrating that VRAIN inserts rare-class objects naturally while preserving annotation quality.

